\documentclass{article}

\PassOptionsToPackage{numbers,sort&compress}{natbib}
\usepackage[dblblindworkshop,final]{neurips_2026}
\workshoptitle{LatinX in AI at NeurIPS 2026}

\usepackage[utf8]{inputenc}
\usepackage[T1]{fontenc}
\usepackage{hyperref}
\usepackage{url}
\usepackage{booktabs}
\usepackage{array}
\usepackage{amsfonts}
\usepackage{amsmath}
\usepackage{amssymb}
\usepackage{amsthm}
\usepackage{microtype}
\usepackage[table]{xcolor}
\usepackage{graphicx}
\usepackage{tikz}
\usepackage{pgfplots}
\usepackage{placeins}
\usetikzlibrary{shapes,arrows.meta,positioning,fit,backgrounds,calc}
\pgfplotsset{compat=1.18}

\newtheorem{theorem}{Theorem}
\newtheorem{corollary}{Corollary}

\title{Protocol-Aware Auditing of Domain-Blind Feature Attribution in Network Intrusion Detection}
\author{%
  Emmanuelli El-Mahmoud \\
  Baylor University, Waco, TX, USA \\
  \texttt{emmanuelli\_elmahmou1@baylor.edu}
  \And
  Abanisenioluwa Kolawole Orojo \\
  Baylor University, Waco, TX, USA \\
  \texttt{Abanisenioluwa\_Oroj1@baylor.edu}
  \AND
  Pablo Rivas \\
  Marist University, Poughkeepsie, NY, USA \\
  \texttt{Pablo.Rivas@marist.edu}
  \And
  Erika Leal \\
  Baylor University, Waco, TX, USA \\
  \texttt{Erika\_Leal@baylor.edu}
}
\hypersetup{
  hidelinks,
  pdftitle={Protocol-Aware Auditing of Domain-Blind Feature Attribution in Network Intrusion Detection},
  pdfauthor={Emmanuelli El-Mahmoud; Abanisenioluwa Kolawole Orojo; Pablo Rivas; Erika Leal},
  pdfsubject={LatinX in AI Research Workshop at NeurIPS 2026},
  pdfkeywords={Explainable AI, Feature Attribution, Network Intrusion Detection, Protocol-Aware XAI}
}

\begin{document}
\maketitle

\begin{abstract}
Feature-attribution methods can query rule-violating network states through perturbations, coalitions, references, paths, and ablations, making explanations and their evaluation sensitive to model behavior outside the operational domain. We formalize this failure mode as domain blindness and prove valid-query invariance: when models agree on the required oracle information within a declared rule-valid set, a procedure confined to that set returns the same attribution. We instantiate one protocol-aware query-control principle in LIME, SHAP, Integrated Gradients, and DeepLIFT and audit paired configurations across four NIDS datasets and three classifier families. In six settings where standard LIME reports perfect top-feature repeatability, PA-LIME reveals substantial variability, with a mean of 0.575. Under FGSM followed by protocol repair, PA-DeepLIFT and PA-IG achieve the largest attribution-similarity curve areas, 0.452 and 0.413, while mean within-PA rank agreement rises to 0.366 from 0.217. PA-XAI therefore turns structured domain assumptions into explicit, testable controls for explanation auditing; support modeling, raw-flow reconstruction, and adaptive explanation attacks provide complementary validation layers.
\end{abstract}

\section{Introduction}
\label{sec:introduction}

Feature attributions help analysts inspect NIDS and other security models
\citep{warnecke2020evaluating,ozawa2025criteria}, yet practitioner studies show
that explanations must support verification, investigation, and downstream
reasoning to earn a place in security workflows
\citep{mink2023everybody,schoenherr2023designing}. We ask whether LIME
neighborhoods, SHAP coalitions, IG paths, DeepLIFT references, and ablation
states respect discrete, protocol-gated, ordered, and derived-feature rules.
Violations create \emph{domain blindness}, even when the defender has direct
model access \citep{aas2021dependent,anders2020fairwashing,zaher2024manifold}.

We define a rule-valid set $\mathcal V_{\mathcal R}$ and an
explainer-oracle transcript. Theorem~\ref{thm:valid-query} shows that a contained
run is invariant when models provide identical required oracle information on
that set. PA-XAI applies one query-control principle across four explainer
families and audits model alignment, attack-followed-by-repair stability,
cross-configuration agreement, and feature-reduction utility. Across four
datasets, six standard-LIME perfect-repeatability cases have a paired PA-LIME
mean of $0.575$; its faithfulness
coefficient also changes sign in $5/12$ comparisons. The remaining modules
reveal distinct method-specific gains and trade-offs, demonstrating why
explanation reliability requires complementary evidence rather than a single
method ranking.

\paragraph{Contributions.}
\textbf{(1) Formal.} We define procedure-level domain blindness through an
explicit rule-valid set and explainer-specific oracle transcript, prove
valid-query invariance, and extend the same identifiability condition to finite
model-query evaluation metrics. \textbf{(2) Method.} We express PA-LIME,
PA-SHAP, PA-IG, and PA-DeepLIFT as four instances of a common query-control
principle, with backend-specific scopes, and connect them through a
six-coordinate audit profile. \textbf{(3) Empirical.} Across four NIDS datasets
and three classifier families, paired PA configurations change LIME faithfulness
conclusions and repeatability, show higher within-PA agreement, and extend the
observed compression--fidelity frontier.
\section{Related Work and Formal Problem}
\label{sec:related-formal}

\subsection{Positioning}
\paragraph{Security and NIDS.}
Security-focused studies evaluate attribution in malware, vulnerability, and
NIDS tasks \citep{warnecke2020evaluating,neupane2022explainable,
tritscher2023evaluating,arreche2024exai,arreche2025whitebox}. CoLIME preserves
pairwise dependence in LIME neighborhoods \citep{bacevicius2026colime}, while
other work connects explanations to threat intelligence and active response
\citep{ozawa2025criteria,wei2023xnids}. Practitioner and human-centered studies
show that trustworthy use also depends on verification, task fit, and broader
workflow requirements \citep{mink2023everybody,schoenherr2023designing};
dataset audits expose benchmark artifacts \citep{flood2024design}. Our audit
extends this line by treating containment within declared encoded-feature rules
as a shared object across four explainer families.

\paragraph{Explanation evaluation.}
Infidelity and sensitivity test model alignment and local variation, whereas
ROAR retrains after feature removal \citep{yeh2019fidelity,hooker2019benchmark}.
Sanity checks test model and label dependence, and encoding explanations can
defeat removal scores \citep{adebayo2018sanity,puli2024encoding}; complete
linear attributions need not recover task behavior
\citep{bilodeau2024impossibility}. LIME studies analyze weighting, sampling,
locality, and gradient links
\citep{garreau2020explaining,tan2023glime,agarwal2021unification}. Models can
also detect and fool LIME/SHAP perturbations \citep{slack2020fooling}, while
external audits may require many queries \citep{bhattacharjee2024auditing}.
Our corollary places explanation generation and finite-query evaluation under a
single transcript-level identifiability condition.

\paragraph{Dependence and valid domains.}
Conditional and on-manifold Shapley methods account for dependence or restrict
imputations to supported states \citep{aas2021dependent,frye2021shapley}.
Off-manifold analysis shows that models can agree on observed data yet differ
in their explanations \citep{anders2020fairwashing}; ManifoldShap restricts
coalitions to a learned manifold \citep{taufiq2023manifold}. Geometry-aware IG,
CHILLI, and representation studies expose related path, perturbation, and
encoding choices \citep{zaher2024manifold,anwar2024chilli,
hwang2025representation}. Complementing these approaches, our framework makes
an explicit structured rule set the common object for neighborhood, coalition,
path/reference, and evaluation queries.

\paragraph{Adversarial settings.}
FGSM and PGD motivate the classifier stress tests
\citep{goodfellow2014explaining,madry2018towards}. Prior work studies
input-level explanation manipulation, joint prediction-and-interpretation
attacks, explanation-aware backdoors, and their threat models
\citep{dombrowski2019manipulated,zhang2020underfire,noppel2023backdoors,
noppel2024sok}. NIDS work also uses XAI-guided problem-space evasion
\citep{okada2025xai}. Our module evaluates the repair pipeline and positions
adaptive constrained and explanation-targeting attacks as complementary threat
models.

\subsection{Formal problem}
\label{sec:formal-problem}
Let $f:\mathcal{X}\rightarrow\mathbb{R}^{K}$ be a classifier on the encoded
space
\begin{equation}
\mathcal{X}=\prod_{i\in\mathcal{I}_{c}}\mathcal{X}_{i}
\times\prod_{g\in\mathcal{G}_{d}}\mathcal{D}_{g},
\label{eq:encoded-space}
\end{equation}
where $\mathcal{I}_{c}$ indexes continuous coordinates and each finite
$\mathcal{D}_{g}$ contains permitted patterns for a discrete feature group. A
finite rule set
$\mathcal{R}=\mathcal{R}_{\mathrm{dom}}\cup\mathcal{R}_{\mathrm{ord}}
\cup\mathcal{R}_{\mathrm{gate}}\cup\mathcal{R}_{\mathrm{der}}$ contains domain,
order, protocol-gate, and derived-feature predicates and defines
\begin{equation}
\mathcal{V}_{\mathcal{R}}=\{x\in\mathcal{X}:r(x)=1
\text{ for every }r\in\mathcal{R}\}.
\label{eq:rule-valid-set}
\end{equation}
Examples include nonnegative durations and allowed one-hot protocols, max
$\geq$ mean $\geq$ min, zero TCP-only flags outside TCP, and a stored range
equal to max minus min.

For an attribution or attribution-evaluation procedure $E$, a request
$q\in\mathcal{Q}_{E}$ has finite state footprint
$\sigma_E(q)\subseteq\mathcal{X}$ and returns method-specific information
$\mathcal{O}^{E}_{f}(q)$. At target $x\in\mathcal{V}_{\mathcal{R}}$ and random
state $\omega$, a finite run has transcript and queried-state set
\begin{equation}
\mathsf{T}_{E}(f,x;\omega)=
\bigl((q_t,\mathcal{O}^{E}_{f}(q_t))\bigr)_{t=1}^{N_E},
\qquad
\mathsf{Q}_{E}(f,x;\omega)=\bigcup_{t=1}^{N_E}\sigma_E(q_t).
\label{eq:query-transcript}
\end{equation}
Equation~\eqref{eq:query-transcript} records the queries and model replies
that determine the explanation.
The oracle may return scores, gradients, activations, or multipliers, covering
LIME, sampled SHAP, IG, DeepLIFT, and removal-based evaluation. A run is
\emph{valid-query contained} when
$\mathsf{Q}_{E}(f,x;\omega)\subseteq\mathcal{V}_{\mathcal{R}}$. Its conceptual
leakage probability is
\begin{equation}
\Lambda_E(f,x)=\Pr_{\omega}\!\left[
\mathsf{Q}_{E}(f,x;\omega)\nsubseteq\mathcal{V}_{\mathcal{R}}\right].
\label{eq:validity-leakage}
\end{equation}
Equation~\eqref{eq:validity-leakage} is the probability that at least one
queried state violates a declared rule.
We use $\Lambda_E$ as a formal diagnostic; empirical estimation requires
logging complete query transcripts. The experiments enforce
rule validity for the named operations; empirical-support consistency and
raw-event reconstructability form progressively stronger validation layers.

For a second classifier $g$, write
$f\equiv_{E,\mathcal{V}_{\mathcal{R}}}g$ when their method-specific oracle
responses agree for every request whose footprint lies in
$\mathcal{V}_{\mathcal{R}}$.
\begin{theorem}[Valid-query invariance]
\label{thm:valid-query}
Fix $x\in\mathcal{V}_{\mathcal{R}}$ and $\omega$. Assume every model-dependent
access is recorded in the transcript and, once $\omega$ is fixed, the next
request, stopping decision, and readout depend only on $x$ and the observed
prefix. If $f\equiv_{E,\mathcal{V}_{\mathcal{R}}}g$ and the run against $f$ is
valid-query contained, then
$\mathsf{T}_{E}(f,x;\omega)=\mathsf{T}_{E}(g,x;\omega)$ and
$E(f,x;\omega)=E(g,x;\omega)$. Conversely, if two backend-admissible oracle
maps differ only at one off-rule-valid request and that reply changes the
output, they agree on every rule-valid request yet give different outputs. A
classifier-level witness follows when classifiers induce both maps.
\end{theorem}
\begin{proof}[Proof sketch]
Couple both runs with the same $x$ and $\omega$. Equal transcript prefixes give
the same next request and stopping decision. Containment places each request
inside $\mathcal{V}_{\mathcal{R}}$, where oracle equivalence gives the same
reply; induction gives identical transcripts and readouts. For the converse,
vary only the influential off-rule-valid reply while holding every rule-valid
request fixed.
\end{proof}
The theorem establishes invariance at the oracle-transcript level and connects
explicit rules to perturbation, coalition, path/reference, and evaluation
queries. Corollary~\ref{cor:evaluation-query} extends the same condition to
evaluation requests; full statements and proofs appear in
Appendix~\ref{app:formal}.
\section{Protocol-Aware XAI Audit Framework}
\label{sec:framework}

\paragraph{One query-control principle.}
Each PA instance modifies the explainer-specific object that determines where
model information is obtained: LIME's neighborhood, SHAP's background or
coalition completion, and the references or paths used by IG and DeepLIFT. The
comparisons evaluate complete query configurations, including sampling and
reference choices. \emph{Standard} and \emph{PA} denote the paired configurations
in this study.

\begin{corollary}[Evaluation-query inheritance]
\label{cor:evaluation-query}
Let $\phi=E(f,x;\omega)$ and let a finite value-query metric have the form
\begin{equation}
M(E,f,x;\omega)=\Psi_M\!\left(
\phi,f_{k_1}(r_1(x,\phi)),\ldots,f_{k_m}(r_m(x,\phi))\right).
\label{eq:evaluation-metric}
\end{equation}
Regard $E$ and the score calls in Equation~\eqref{eq:evaluation-metric} as one
oracle procedure. If every explanation and metric query is contained in
$\mathcal{V}_{\mathcal{R}}$ and two models return the same requested oracle
information there, Theorem~\ref{thm:valid-query} gives equal metric values. An
influential off-rule-valid metric query makes the metric non-identifiable from
rule-valid oracle behavior alone.
\end{corollary}
The corollary places explanation generation and finite-query evaluation under
the same identifiability condition, making replacement-state validity part of
the metric's scientific specification. Appendix~\ref{app:evaluation-corollary}
gives the full proof. 

\begin{figure}[t!]
  \centering
  \includegraphics[width=\textwidth]{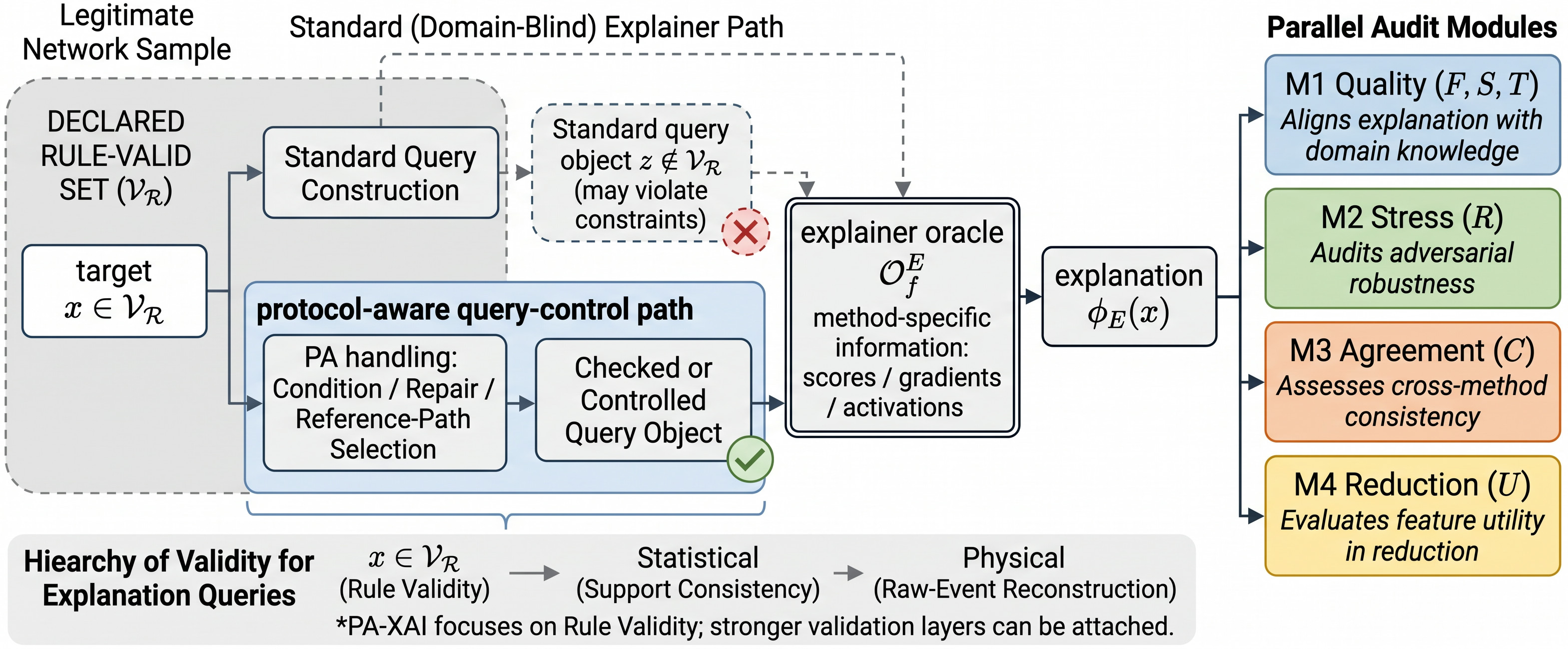}
  \caption{
    Domain assumptions and audit flow. A rule-valid target can still be
    explained through method-specific queries that leave the declared set
    $\mathcal V_{\mathcal R}$. PA-XAI controls the query object before the
    model oracle is evaluated; the resulting explanation is then examined by
    four parallel audit modules. The control is backend-specific;
    empirical-support and raw-event reconstruction tests extend the
    rule-valid audit with progressively stronger validation layers.
  }
  \label{fig:framework-audit}
\end{figure}

Figure~\ref{fig:framework-audit} connects the formal query-level analysis to
the empirical audit: PA-XAI controls the method-specific query object before
model evaluation, and four parallel modules assess the resulting explanation.

\begin{table}[h!]
\centering
\setlength{\tabcolsep}{3pt}
\renewcommand{\arraystretch}{1.08}
\caption{One query-control principle with four backend-specific instances.
$\Pi_{\mathcal R}$ is the ordered rule-repair routine. Appendix~\ref{app:methods}
gives complete procedures and configuration details.}
\label{tab:pa-methods}
\begin{tabular}{@{}>{\raggedright\arraybackslash}p{0.105\linewidth}>{\raggedright\arraybackslash}p{0.155\linewidth}>{\raggedright\arraybackslash}p{0.265\linewidth}>{\raggedright\arraybackslash}p{0.195\linewidth}>{\raggedright\arraybackslash}p{0.215\linewidth}@{}}
\toprule
Method & Standard choice & PA query control & Controlled property & Scope of control \\
\midrule
LIME & Gaussian neighborhood & Sample protocol; draw empirical discrete marginals; perturb continuous coordinates; apply $\Pi_{\mathcal R}$; fit the same surrogate & Repaired neighborhood rows satisfy implemented checks & Joint-support and reconstruction tests are separate validation stages \\
SHAP & Background and masked completion & Use benign backgrounds; repair KernelSHAP completions; pass the background to TreeSHAP/DeepSHAP & KernelSHAP evaluates repaired coalition completions & Tree and gradient backends use the background without repair \\
IG & Zero reference and direct interpolation in the study comparator & Select a threshold-near training reference; integrate continuous coordinates with discrete groups fixed; add ordered endpoint differences & Avoids fractional discrete interpolation and preserves the stated sum identity & Discrete endpoint control alone does not certify full-path containment \\
DeepLIFT & Zero reference in the study comparator & Use the threshold-near training reference with Rescale endpoint differences & Uses an observed reference without interpolated discrete inputs & An observed reference does not certify every internal query \\
\bottomrule
\end{tabular}
\end{table}

Table~\ref{tab:pa-methods} shows that explicit generated-state repair applies
to PA-LIME neighborhoods and KernelSHAP coalition completions, whereas PA-IG
and PA-DeepLIFT control reference and path semantics.
Appendix~\ref{app:methods} details each procedure.

\paragraph{Diagnostic audit profile.}
For fixed audit records $\mathcal{D}$, define
\begin{equation}
\mathbf{A}(E,f,\mathcal{D})=(F,S,R,C,U,T),
\label{eq:audit-profile}
\end{equation}
where $(F,S,R,C,U,T)$ record model alignment, repeated-explanation stability,
attack-followed-by-repair attribution stability, cross-method agreement,
feature-reduction utility, and cost. Modules~1--4 contribute $(F,S,T)$, $R$,
$C$, and $U$, respectively. The six coordinates preserve distinct evidence
types, allowing a reader to see which prerequisites for reliance are supported
for a particular model--explainer configuration.
\section{Experimental Setup}
\label{sec:setup}

\paragraph{Data and preprocessing.}
We evaluate four tabular NIDS benchmarks: NSL-KDD, CIC-IDS-2017, UNSW-NB15,
and CSE-CIC-IDS2018. Median imputation, one-hot encoding, and min--max scaling
precede classification. Training rows define the PA-LIME sampling distributions,
PA-SHAP backgrounds, and PA-IG/PA-DeepLIFT references; Module~4 uses a separate
calibration set and held-out test partition.

\paragraph{Models and explanations.}
The primary analyses use a fully connected DNN, Random Forest (RF), and XGBoost
(XGB). Module~1 pairs SHAP and LIME with all three families and pairs IG and
DeepLIFT with the DNN; every displayed standard configuration has a PA
counterpart. Module~2 attacks the DNN. Module~4 uses a separate RF pruning proxy
and retrains a DNN, 1D-CNN, RF, and SVM on selected features.

\paragraph{Module protocols.}
Module~1 reports Pearson faithfulness after top-ranked mean replacement,
normalized-entropy complexity, sparsity, repeated-run top-$k$ Jaccard, and
per-sample latency. Module~2 evaluates FGSM at
$\varepsilon\!\in\!\{.01,.05,.1,.2,.3\}$ and PGD at
$\varepsilon\!\in\!\{.01,.05,.1\}$, comparing unrestricted attack with attack
followed by rule repair. Module~3 uses $10{,}000$ samples and 66 pairs per
dataset for rank and top-$\{5,10\}$ agreement, plus $5{,}000$ samples per
dataset for SHAP interactions. Module~4 ranks mean absolute attributions over
$5{,}000$ calibration samples, prunes with a 50-tree, depth-10 RF trained on a
$50{,}000$-sample subsample, stops when macro-F1 falls by more than $.01$, and
compares four downstream classifiers with Chi-Squared, Information Gain, PCA,
and Spearman selection.

Appendix~\ref{app:constraints} catalogs the encoded-space rules.
Appendix~\ref{app:setup} consolidates the reported configuration and identifies
the additional dataset, model, software, and execution details required for an
exact rerun. Appendix~\ref{app:numerical} derives the aggregate summaries used
in the main paper.
\section{Results}
\label{sec:results}

\subsection{Module 1: Explanation Quality}
\label{sec:results-quality}

\begin{figure}[b!]
  \centering
\begin{minipage}[t]{0.487\linewidth}
\centering
\begin{tikzpicture}
\begin{axis}[
  width=\linewidth,
  height=4.40cm,
  xmin=0.45,xmax=12.55,
  ymin=0.15,ymax=1.06,
  xtick={1,...,12},
  xticklabels={D-N,D-17,D-U,D-18,R-N,R-17,R-U,R-18,X-N,X-17,X-U,X-18},
  ytick={0.2,0.4,0.6,0.8,1.0},
  xlabel={model--dataset case},
  ylabel={Top-$k$ Jaccard},
  title={(a) Repeated-run stability (12 cases)},
  axis x line*=bottom,
  axis y line*=left,
  tick align=outside,
  xticklabel style={font=\tiny,rotate=55,anchor=east},
  yticklabel style={font=\scriptsize},
  xlabel style={font=\scriptsize},
  ylabel style={font=\scriptsize},
  title style={font=\scriptsize},
  grid=major,
  grid style={black!8},
  clip=false,
]
\addplot[black!35,densely dashed,thin,forget plot] coordinates {(4.5,0.15) (4.5,1.06)};
\addplot[black!35,densely dashed,thin,forget plot] coordinates {(8.5,0.15) (8.5,1.06)};

\addplot[black!22,thin,forget plot] coordinates {(1,0.17) (1,0.52)};
\addplot[black!22,thin,forget plot] coordinates {(2,0.37) (2,0.45)};
\addplot[black!80,semithick,forget plot] coordinates {(3,1) (3,0.48)};
\addplot[black!80,semithick,forget plot] coordinates {(4,1) (4,0.39)};
\addplot[black!22,thin,forget plot] coordinates {(5,0.23) (5,0.74)};
\addplot[black!22,thin,forget plot] coordinates {(6,0.52) (6,0.67)};
\addplot[black!80,semithick,forget plot] coordinates {(7,1) (7,0.69)};
\addplot[black!80,semithick,forget plot] coordinates {(8,1) (8,0.86)};
\addplot[black!22,thin,forget plot] coordinates {(9,0.15) (9,0.62)};
\addplot[black!22,thin,forget plot] coordinates {(10,0.44) (10,0.5)};
\addplot[black!80,semithick,forget plot] coordinates {(11,1) (11,0.43)};
\addplot[black!80,semithick,forget plot] coordinates {(12,1) (12,0.6)};
\addplot[only marks,mark=o,mark size=2.05pt,very thick,black,
  mark options={fill=white}] coordinates {(1,0.17) (2,0.37) (3,1) (4,1) (5,0.23) (6,0.52) (7,1) (8,1) (9,0.15) (10,0.44) (11,1) (12,1)};
\addplot[only marks,mark=square*,mark size=1.95pt,black]
  coordinates {(1,0.52) (2,0.45) (3,0.48) (4,0.39) (5,0.74) (6,0.67) (7,0.69) (8,0.86) (9,0.62) (10,0.5) (11,0.43) (12,0.6)};
\end{axis}
\end{tikzpicture}
\end{minipage}\hfill
\begin{minipage}[t]{0.487\linewidth}
\centering
\begin{tikzpicture}
\begin{axis}[
  width=\linewidth,
  height=4.40cm,
  xmin=0.45,xmax=12.55,
  ymin=-0.35,ymax=0.56,
  xtick={1,...,12},
  xticklabels={D-N,D-17,D-U,D-18,R-N,R-17,R-U,R-18,X-N,X-17,X-U,X-18},
  ytick={-0.3,0,0.3,0.5},
  xlabel={model--dataset case},
  ylabel={Reported faithfulness $r$},
  title={(b) Mean-replacement faithfulness (12 cases)},
  axis x line*=bottom,
  axis y line*=left,
  tick align=outside,
  xticklabel style={font=\tiny,rotate=55,anchor=east},
  yticklabel style={font=\scriptsize},
  xlabel style={font=\scriptsize},
  ylabel style={font=\scriptsize},
  title style={font=\scriptsize},
  grid=major,
  grid style={black!8},
  clip=false,
]
\addplot[black!35,densely dashed,thin,forget plot] coordinates {(4.5,-0.35) (4.5,0.56)};
\addplot[black!35,densely dashed,thin,forget plot] coordinates {(8.5,-0.35) (8.5,0.56)};
\addplot[black!55,densely dashed,thin,forget plot] coordinates {(0.5,0) (12.5,0)};
\addplot[black!80,semithick,forget plot] coordinates {(1,-0.05) (1,0.2)};
\addplot[black!80,semithick,forget plot] coordinates {(2,0.5) (2,-0.05)};
\addplot[black!80,semithick,forget plot] coordinates {(3,0.19) (3,-0.14)};
\addplot[black!80,semithick,forget plot] coordinates {(4,0.49) (4,-0.3)};
\addplot[black!22,thin,forget plot] coordinates {(5,0.18) (5,0.66)};
\addplot[black!22,thin,forget plot] coordinates {(6,0.31) (6,0.39)};
\addplot[black!22,thin,forget plot] coordinates {(7,0.66) (7,0.56)};
\addplot[black!22,thin,forget plot] coordinates {(8,0.36) (8,0.36)};
\addplot[black!22,thin,forget plot] coordinates {(9,0.14) (9,0.38)};
\addplot[black!22,thin,forget plot] coordinates {(10,0.21) (10,0.29)};
\addplot[black!22,thin,forget plot] coordinates {(11,0.25) (11,0.1)};
\addplot[black!80,semithick,forget plot] coordinates {(12,0.31) (12,-0.07)};
\addplot[only marks,mark=o,mark size=2.05pt,very thick,black,
  mark options={fill=white}] coordinates {(1,-0.05) (2,0.5) (3,0.19) (4,0.49) (5,0.18) (6,0.31) (7,0.66) (8,0.36) (9,0.14) (10,0.21) (11,0.25) (12,0.31)};
\addplot[only marks,mark=square*,mark size=1.95pt,black]
  coordinates {(1,0.2) (2,-0.05) (3,-0.14) (4,-0.3) (5,0.66) (6,0.39) (7,0.56) (8,0.36) (9,0.38) (10,0.29) (11,0.1) (12,-0.07)};
\end{axis}
\end{tikzpicture}
\end{minipage}
  \caption{Paired standard-LIME (open circles) and PA-LIME (filled squares)
  results for all 12 model--dataset cases. D, R, and X denote DNN, RF, and
  XGBoost; N, 17, U, and 18 denote NSL-KDD, CIC-IDS-2017, UNSW-NB15, and
  CSE-CIC-IDS2018. (a) Top-$k$ repeated-run Jaccard; dark connectors mark the
  six cases with standard $J=1.0$. (b) Reported Pearson mean-replacement
  faithfulness; dark connectors mark sign changes and the dashed line marks
  zero. Higher values indicate stronger alignment under each stated metric.}
  \label{fig:module1-quality}
\end{figure}
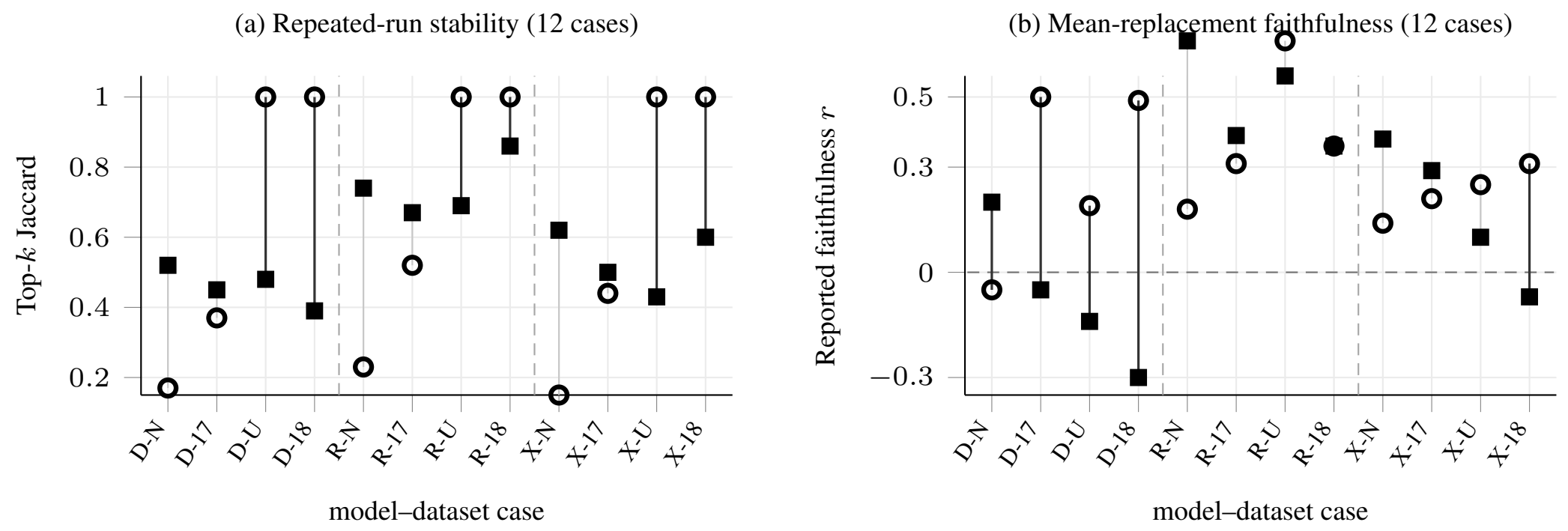

\paragraph{Query-sensitive repeatability.}
Across the 12 LIME model--dataset comparisons, standard LIME reports perfect
top-$k$ repeated-run overlap in six settings: all three classifier families on
UNSW-NB15 and CSE-CIC-IDS2018. The paired PA-LIME values are $0.48$, $0.39$,
$0.69$, $0.86$, $0.43$, and $0.60$, with mean $0.575$ and an absolute
difference of $-0.425$ from the standard value of $1.0$
(Figure~\ref{fig:module1-quality}a). The decrease from perfect overlap to a mean
of $0.575$ shows that repeatability depends on neighborhood construction.
PA-LIME changes sampling and repair together; the comparison measures their
combined effect.

\paragraph{Faithfulness conclusions and method ranking.}
The reported mean-replacement correlation changes sign in five of 12 LIME
comparisons, including all four DNN comparisons
(Figure~\ref{fig:module1-quality}b). Averaged across the four datasets, the DNN
ranking changes from SHAP ($0.63$) $>$ IG ($0.57$) $>$ DeepLIFT ($0.51$) $>$
LIME ($0.28$) under the standard configurations to SHAP ($0.62$) $>$ DeepLIFT
($0.54$) $>$ IG ($0.43$) $>$ LIME ($-0.07$) under the PA configurations. The
domain-sensitive query choice therefore changes which explainer the metric
selects. A negative coefficient denotes inverse association under this
ablation protocol, making the shift a substantive change in the evaluator's
conclusion.

\paragraph{Repeatability and cost.}
The displayed SHAP, IG, and DeepLIFT configurations report $J=1.0$ under both
settings, confirming exact repeated-run identity for the fixed deterministic
computations. The timing profile separates lower-latency gradient/reference
methods from perturbation-based auditing: the reported PA-DeepLIFT and PA-IG
means are $6.7$ and $54.6$ ms per explanation, while PA-LIME spans
$1690.6$--$1738.0$ ms and PA-SHAP spans $248.5$--$1990.7$ ms. Appendix~\ref{app:quality}
provides the complete values and metric definitions.

\subsection{Module 2: Attack Followed by Protocol Repair}
\label{sec:results-adversarial}
Let $A_{\varepsilon}$ be the reported FGSM or PGD classifier-evasion routine
and $\Pi_{\mathcal R}$ the implemented rule-repair map. The evaluated pipeline
and its adaptive valid-set extension are
\begin{align}
 x_{\mathrm{atk}}&=A_{\varepsilon}(f,x,y),\quad
 x_{\mathrm{rep}}=\Pi_{\mathcal R}(x_{\mathrm{atk}}),
 \label{eq:attack-repair}\\
 x^{\star}_{\mathrm{ad}}&\in
 \arg\max_{\substack{z\in\mathcal V_{\mathcal R}\\
 \rho(z,x)\leq\varepsilon}}
 \mathcal L_{\mathrm{cls}}(f(z),y).
 \label{eq:adaptive-constrained-attack}
\end{align}
Equation~\eqref{eq:attack-repair} defines the evaluated repair pipeline;
Equation~\eqref{eq:adaptive-constrained-attack} defines adaptive valid-set
optimization, which is reserved for future work.

\paragraph{Classifier evasion.}
At $\varepsilon=0.1$, repair changes FGSM success from $63.7\%$ to $51.5\%$
($-12.2$ percentage points) and PGD from $74.6\%$ to $58.8\%$ ($-15.8$
points). Across the evaluated budgets, normalized trapezoidal area changes from
$64.09$ to $57.86$ for FGSM and from $52.86$ to $44.99$ for PGD, descriptive
reductions of $9.7\%$ and $14.9\%$ (See Figure~\ref{fig:appendix-module2-summary}a in Appendix~\ref{app:adversarial}).
These reductions quantify the effect of the repair pipeline on the tested
classifier-evasion attacks.

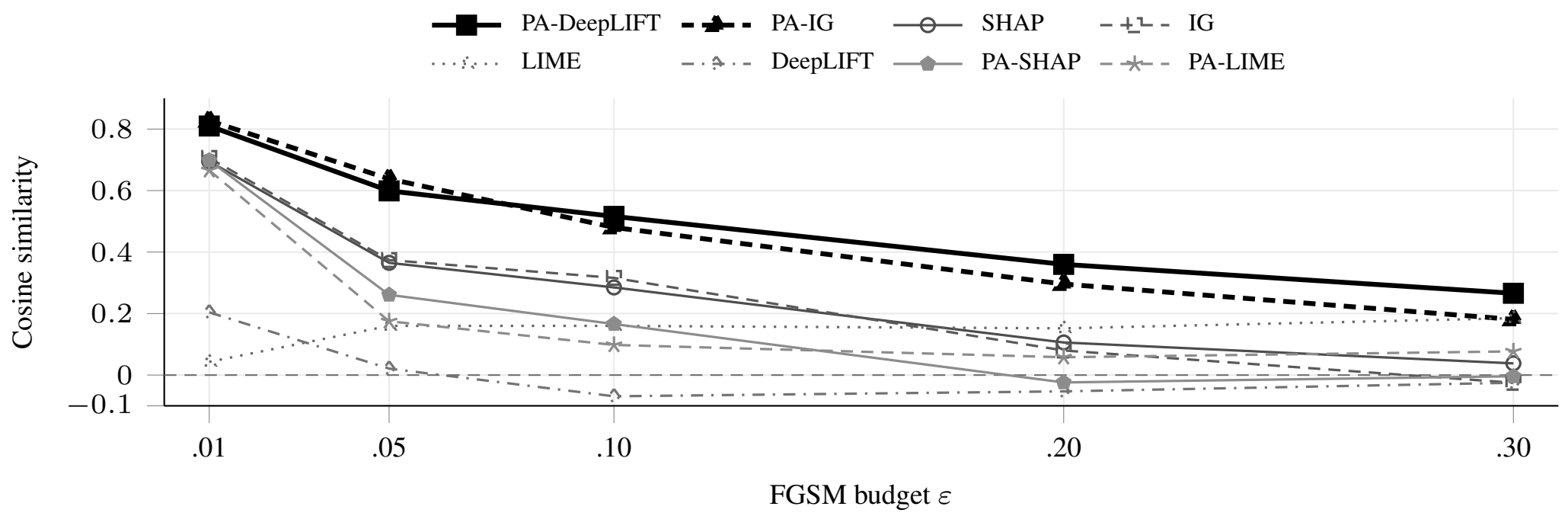
\begin{figure}[b!]
  \centering
\begin{tikzpicture}
\begin{axis}[
  width=0.98\linewidth,
  height=4.25cm,
  xmin=0.0,xmax=0.31,
  ymin=-0.10,ymax=0.90,
  xtick={0.01,0.05,0.10,0.20,0.30},
  xticklabels={.01,.05,.10,.20,.30},
  ytick={-0.1,0,0.2,0.4,0.6,0.8},
  xlabel={FGSM budget $\varepsilon$},
  ylabel={Cosine similarity},
  axis x line*=bottom,
  axis y line*=left,
  tick align=outside,
  tick label style={font=\scriptsize},
  xlabel style={font=\scriptsize},
  ylabel style={font=\scriptsize},
  grid=major,
  grid style={black!8},
  legend style={font=\tiny,at={(0.5,1.02)},anchor=south,draw=none,legend columns=4,column sep=2.5pt,row sep=0pt},
  legend cell align={left},
  clip=false,
]
\addplot[black!55,densely dashed,thin,forget plot] coordinates {(0,0) (0.31,0)};
\addplot[black,very thick,solid,mark=square*,mark size=2.0pt] coordinates {(0.01,0.81) (0.05,0.599) (0.1,0.516) (0.2,0.36) (0.3,0.266)};
\addlegendentry{PA-DeepLIFT}
\addplot[black,very thick,densely dashed,mark=triangle*,mark size=2.2pt] coordinates {(0.01,0.827) (0.05,0.638) (0.1,0.48) (0.2,0.296) (0.3,0.181)};
\addlegendentry{PA-IG}
\addplot[black!70,semithick,solid,mark=o,mark size=1.7pt,mark options={fill=white}] coordinates {(0.01,0.695) (0.05,0.365) (0.1,0.285) (0.2,0.106) (0.3,0.038)};
\addlegendentry{SHAP}
\addplot[black!65,semithick,densely dashed,mark=square,mark size=1.7pt,mark options={fill=white}] coordinates {(0.01,0.706) (0.05,0.374) (0.1,0.316) (0.2,0.08) (0.3,-0.025)};
\addlegendentry{IG}
\addplot[black!60,semithick,dotted,mark=triangle,mark size=1.8pt,mark options={fill=white}] coordinates {(0.01,0.043) (0.05,0.16) (0.1,0.16) (0.2,0.152) (0.3,0.185)};
\addlegendentry{LIME}
\addplot[black!55,semithick,dashdotted,mark=diamond,mark size=1.8pt,mark options={fill=white}] coordinates {(0.01,0.204) (0.05,0.021) (0.1,-0.069) (0.2,-0.053) (0.3,-0.025)};
\addlegendentry{DeepLIFT}
\addplot[black!45,semithick,solid,mark=pentagon*,mark size=1.7pt] coordinates {(0.01,0.698) (0.05,0.261) (0.1,0.166) (0.2,-0.024) (0.3,-0.004)};
\addlegendentry{PA-SHAP}
\addplot[black!45,semithick,densely dashed,mark=star,mark size=2.1pt] coordinates {(0.01,0.666) (0.05,0.175) (0.1,0.098) (0.2,0.058) (0.3,0.077)};
\addlegendentry{PA-LIME}
\end{axis}
\end{tikzpicture}
  \caption{Mean cosine similarity under FGSM followed by repair. All 8
  methods and 5 budgets shown; thick lines emphasize PA-DeepLIFT
  and PA-IG. Higher values $=$ greater directional alignment.}
  \label{fig:module2-adversarial}
\end{figure}

\paragraph{Attribution stability.}
PA-DeepLIFT and PA-IG rank first and second by cosine nAUC, at $0.452$ and
$0.413$ (Figure~\ref{fig:module2-adversarial}). The ordering remains
method-specific: at $\varepsilon=0.3$, LIME ($0.185$) slightly exceeds PA-IG
($0.181$), and the PA-LIME and PA-SHAP curve summaries trail their standard
counterparts.

\paragraph{Perturbed-feature localization.}
Relative to LIME, PA-LIME raises Top-1 recovery from $0.064$ to $0.152$
($2.38\times$), Top-3 recovery from $0.200$ to $0.384$ ($1.92\times$), and
reduces mean perturbed-feature rank from $73.0$ to $41.2$ ($43.6\%$). These
gains identify perturbation-sensitive features for further investigation;
analyst benefit remains a separate evaluation target. Appendix~\ref{app:adversarial} gives the complete curves,
localization table, and threat-model interpretation.

\subsection{Module 3: Agreement, Interactions, and Reference Sets}
\label{sec:results-consensus}
\newcommand{\MainConsensusTable}{%
\begin{table}[!h]
  \caption{Agreement over 66 configuration pairs per dataset.
  $\bar\rho_{\rm all}$ and $J_5$ are all-pair means; the last two columns are
  within-mode Spearman means. N, C17, U, and C18 denote the four datasets.}
  \label{tab:module3-agreement}
  \centering
  \footnotesize
  \setlength{\tabcolsep}{4.5pt}
  \begin{tabular}{@{}lrrrr@{}}
    \toprule
    Data & $\bar\rho_{\rm all}$ & $J_5$ & $\bar\rho_{\rm std}$ & $\bar\rho_{\rm PA}$ \\
    \midrule
    N & 0.262 & 0.226 & 0.163 & 0.366 \\
    C17 & 0.256 & 0.168 & 0.244 & 0.378 \\
    U & 0.309 & 0.322 & 0.231 & 0.414 \\
    C18 & 0.256 & 0.213 & 0.230 & 0.306 \\
    \bottomrule
  \end{tabular}
\end{table}%
}

\newcommand{\FullConsensusTable}{%
\begin{table}[!h]
  \caption{Aggregate agreement records. Each row summarizes 66 configuration
  pairs; $J_5$ and $J_{10}$ are feature-set overlaps. Wilcoxon is the reported
  percentage of pairings rejecting attribution equivalence at nominal
  $\alpha=0.05$.}
  \label{tab:consensus-full}
  \centering
  \scriptsize
  \setlength{\tabcolsep}{4.0pt}
  \begin{tabular}{@{}lcccccccc@{}}
    \toprule
    Dataset & $\rho_s$ & $\tau_b$ & $J_5$ & $J_{10}$ & Wilc. (\%) & $\rho_{\rm std}$ & $\rho_{\rm PA}$ \\
    \midrule
    NSL-KDD & $0.262\pm0.259$ & $0.220\pm0.229$ & 0.226 & 0.300 & 84.8 & 0.163 & 0.366 \\
    CIC-IDS-2017 & $0.256\pm0.207$ & $0.196\pm0.169$ & 0.168 & 0.240 & 89.4 & 0.244 & 0.378 \\
    UNSW-NB15 & $0.309\pm0.132$ & $0.249\pm0.110$ & 0.322 & 0.328 & 84.8 & 0.231 & 0.414 \\
    CSE-CIC-IDS2018 & $0.256\pm0.122$ & $0.206\pm0.098$ & 0.213 & 0.262 & 84.8 & 0.230 & 0.306 \\
    \bottomrule
  \end{tabular}
\end{table}%
}

\newcommand{\PairExamplesTable}{%
\begin{table}[!h]
  \caption{Selected pair examples. The context column identifies which factors
  vary; values marked $>$ are reported thresholds rather than exact pair means.}
  \label{tab:consensus-pair-examples}
  \centering
  \scriptsize
  \setlength{\tabcolsep}{4.2pt}
  \begin{tabular}{@{}llcl@{}}
    \toprule
    Dataset & Pair & $\rho_s$ & Comparison context \\
    \midrule
    NSL-KDD & DNN-LIME vs. DNN-SHAP & $-0.137$ & Fixed DNN; explainer changes \\
    NSL-KDD & PA-DeepLIFT vs. PA-IG & $>0.91$ & DNN-only; shared reference rule \\
    CIC-IDS-2017 & PA-DeepLIFT vs. PA-IG & $>0.91$ & DNN-only; shared reference rule \\
    CSE-CIC-IDS2018 & PA-DeepLIFT vs. PA-IG & $0.971$ & DNN-only; shared reference rule \\
    CSE-CIC-IDS2018 & DNN-PA-LIME vs. RF-LIME & $-0.239$ & Model and generation mode change \\
    \bottomrule
  \end{tabular}
\end{table}%
}

\newcommand{\InteractionTable}{%
\begin{table}[!h]
  \caption{Model-specific interaction statistics and their displayed DNN/RF
  quotients. DNN/RF divides the first statistic by the second; their
  aggregation and normalization are unspecified. The final columns report the
  LIME--interaction association on the DNN and its $p$-value.}
  \label{tab:interaction-safe}
  \centering
  \scriptsize
  \setlength{\tabcolsep}{5.0pt}
  \begin{tabular}{@{}lrrrrr@{}}
    \toprule
    Dataset & DNN stat. & RF stat. & DNN/RF & LIME $r$ & $p$ \\
    \midrule
    NSL-KDD & 2.10 & 1.31 & 1.60 & -0.535 & $2.1\times10^{-10}$ \\
    CIC-IDS-2017 & 3.06 & 1.95 & 1.57 & 0.415 & $1.4\times10^{-4}$ \\
    UNSW-NB15 & 1.50 & 1.47 & 1.02 & 0.483 & $8.8\times10^{-6}$ \\
    CSE-CIC-IDS2018 & 3.30 & 2.24 & 1.47 & 0.280 & $1.3\times10^{-2}$ \\
    \bottomrule
  \end{tabular}
\end{table}%
}

\newcommand{\ReferenceSetTable}{%
\begin{table}[!h]
  \caption{CSE-CIC-IDS2018 RRA@15 results against predefined domain reference
  sets. Seven rows are reported attack-specific leaders; the PA-LIME row is an
  all-attack average.}
  \label{tab:expert-reference-selected}
  \centering
  \scriptsize
  \setlength{\tabcolsep}{5.0pt}
  \begin{tabular}{@{}lllrl@{}}
    \toprule
    Attack & Model & Method & RRA@15 & Record type \\
    \midrule
    Bot & DNN & DeepLIFT & 0.589 & attack-specific leader \\
    FTP-BruteForce & DNN & DeepLIFT & 0.404 & attack-specific leader \\
    DoS GoldenEye & DNN & PA-IG & 0.495 & attack-specific leader \\
    DoS Hulk & DNN & PA-IG & 0.464 & attack-specific leader \\
    Infiltration & DNN & PA-IG & 0.471 & attack-specific leader \\
    SSH-Bruteforce & RF & PA-SHAP & 0.631 & attack-specific leader \\
    DDoS-LOIC-HTTP & RF & PA-SHAP & 0.563 & attack-specific leader \\
    All attacks (mean) & DNN & PA-LIME & 0.064 & reported average \\
    \bottomrule
  \end{tabular}
\end{table}%
}

\paragraph{Configuration agreement and within-PA coherence.}
Across 66 configuration pairs per dataset, all-pair mean Spearman agreement
ranges from $0.256$ to $0.309$, while Top-5 overlap ranges from $0.168$ to
$0.322$ (Table~\ref{tab:module3-agreement}). Within-PA Spearman means exceed
within-standard means on every dataset; their four-dataset averages are $0.366$
and $0.217$, an absolute difference of $0.149$. The result reveals stronger
coherence within the PA configurations while preserving substantial
configuration-level variability.

\MainConsensusTable

\paragraph{Model and explainer contributions.}
The aggregate pool combines predictive models and explanation procedures. A
fixed-DNN comparison, DNN-LIME versus DNN-SHAP on NSL-KDD, gives
$\rho=-0.137$ and isolates explainer disagreement. Cross-model pairs capture
the combined influence of model and explanation choice. PA-DeepLIFT and PA-IG
exceed $0.91$ on NSL-KDD and CIC-IDS-2017 and reach $0.971$ on
CSE-CIC-IDS2018; their shared DNN and threshold-near reference rule clarify how common
reference semantics contribute to the observed coherence.

\paragraph{Interactions and predefined domain sets.}
The reported DNN interaction statistic exceeds the RF statistic on all four
datasets, with displayed-value quotients of $1.02$ to $1.60$ and selected
rank gaps of $70$--$71$ positions. Because its aggregation and normalization
are unspecified, this statistic supports only a descriptive comparison. On
CSE-CIC-IDS2018, RRA@15 identifies attack-dependent leaders across predefined
6--8-feature domain sets, showing that correspondence between attributions and
domain signatures is configuration-specific. Agreement and domain-set
alignment complement Module~1. RRA@15 evaluates attack-domain feature sets,
whereas Module~2 localization targets coordinates changed by a perturbation
(Appendix~\ref{app:reference-sets}).

\subsection{Module 4: Attribution-Guided Feature Reduction}
\label{sec:results-reduction}
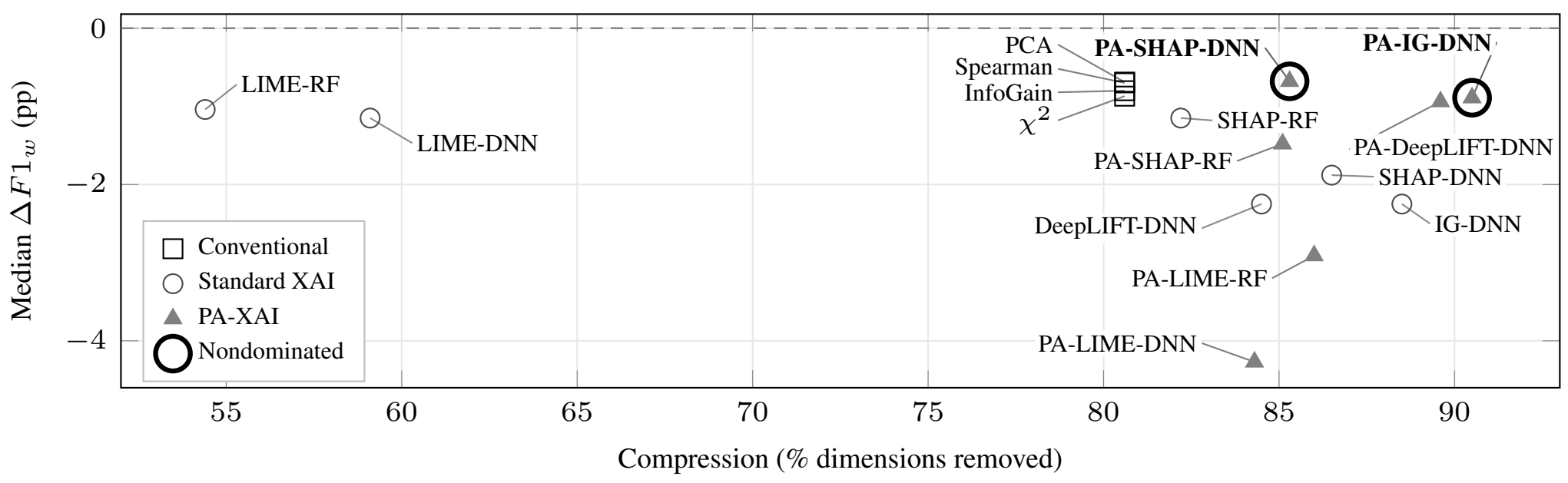
\begin{figure}[t]
  \centering

\begin{tikzpicture}
\begin{axis}[
  width=\linewidth,
  height=4.8cm,
  xmin=52,
  xmax=93.0,
  ymin=-4.6,
  ymax=0.18,
  xlabel={Compression (\% dimensions removed)},
  ylabel={Median $\Delta F1_w$ (pp)},
  tick label style={font=\scriptsize},
  label style={font=\scriptsize},
  grid=major,
  grid style={black!9},
  clip=false,
  legend style={
    font=\tiny,
    at={(0.015,0.015)},
    anchor=south west,
    draw=black!25,
    fill=white,
    row sep=-1pt,
    cells={anchor=west}
  },
  legend cell align={left},
]


\addplot[
  black!55,
  densely dashed,
  thin,
  forget plot
] coordinates {
  (52,0)
  (93,0)
};


\addplot[
  only marks,
  mark=square,
  mark size=2.35pt,
  black,
  mark options={fill=white}
] coordinates {
  (80.6,-0.87) 
  (80.6,-0.69) 
  (80.6,-0.70) 
  (80.6,-0.80) 
};
\addlegendentry{Conventional}


\addplot[
  only marks,
  mark=o,
  mark size=2.35pt,
  black!70,
  mark options={fill=white}
] coordinates {
  (86.5,-1.88) 
  (82.2,-1.15) 
  (59.1,-1.15) 
  (54.4,-1.04) 
  (88.5,-2.25) 
  (84.5,-2.25) 
};
\addlegendentry{Standard XAI}


\addplot[
  only marks,
  mark=triangle*,
  mark size=2.35pt,
  black!50
] coordinates {
  (85.3,-0.68) 
  (85.1,-1.48) 
  (84.3,-4.27) 
  (86.0,-2.91) 
  (90.5,-0.89) 
  (89.6,-0.94) 
};
\addlegendentry{PA-XAI}


\addplot[
  only marks,
  mark=o,
  mark size=4.25pt,
  very thick,
  black,
  mark options={fill=none}
] coordinates {
  (85.3,-0.68) 
  (90.5,-0.89) 
};
\addlegendentry{Nondominated}


\tikzset{
  ann/.style={
    font=\tiny,
    fill=white,
    fill opacity=0.92,
    text opacity=1,
    inner sep=1pt
  },
  annb/.style={
    ann,
    font=\tiny\bfseries
  }
}



\draw[black!50,thin]
  (axis cs:80.6,-0.69) -- (axis cs:78.7,-0.21);
\node[ann,anchor=east] at (axis cs:78.7,-0.21) {PCA};

\draw[black!50,thin]
  (axis cs:80.6,-0.70) -- (axis cs:78.7,-0.52);
\node[ann,anchor=east] at (axis cs:78.7,-0.52) {Spearman};

\draw[black!50,thin]
  (axis cs:80.6,-0.80) -- (axis cs:78.7,-0.82);
\node[ann,anchor=east] at (axis cs:78.7,-0.82) {InfoGain};

\draw[black!50,thin]
  (axis cs:80.6,-0.87) -- (axis cs:78.7,-1.18);
\node[ann,anchor=east] at (axis cs:78.7,-1.18) {$\chi^2$};


\draw[black!45,thin]
  (axis cs:54.4,-1.04) -- (axis cs:55.3,-0.72);
\node[ann,anchor=west] at (axis cs:55.3,-0.72) {LIME-RF};

\draw[black!45,thin]
  (axis cs:59.1,-1.15) -- (axis cs:60.3,-1.47);
\node[ann,anchor=west] at (axis cs:60.3,-1.47) {LIME-DNN};

\draw[black!45,thin]
  (axis cs:82.2,-1.15) -- (axis cs:83.1,-1.18);
\node[ann,anchor=west] at (axis cs:83.1,-1.18) {SHAP-RF};

\draw[black!45,thin]
  (axis cs:86.5,-1.88) -- (axis cs:87.7,-1.9);
\node[ann,anchor=west] at (axis cs:87.7,-1.9) {SHAP-DNN};

\draw[black!45,thin]
  (axis cs:84.5,-2.25) -- (axis cs:82.8,-2.56);
\node[ann,anchor=east] at (axis cs:82.8,-2.56) {DeepLIFT-DNN};

\draw[black!45,thin]
  (axis cs:88.5,-2.25) -- (axis cs:89.3,-2.50);
\node[ann,anchor=west] at (axis cs:89.3,-2.50) {IG-DNN};


\draw[black!60,thin]
  (axis cs:85.3,-0.68) -- (axis cs:84.6,-0.25);
\node[annb,anchor=east] at (axis cs:84.6,-0.25) {PA-SHAP-DNN};

\draw[black!50,thin]
  (axis cs:85.1,-1.48) -- (axis cs:83.8,-1.70);
\node[ann,anchor=east] at (axis cs:83.8,-1.70) {PA-SHAP-RF};

\draw[black!50,thin]
  (axis cs:84.3,-4.27) -- (axis cs:82.8,-4.03);
\node[ann,anchor=east] at (axis cs:82.8,-4.03) {PA-LIME-DNN};

\draw[black!50,thin]
  (axis cs:86.0,-2.91) -- (axis cs:84.8,-3.20);
\node[ann,anchor=east] at (axis cs:84.8,-3.20) {PA-LIME-RF};

\draw[black!60,thin]
  (axis cs:90.5,-0.89) -- (axis cs:91.2,-0.18);
\node[annb,anchor=east] at (axis cs:91.2,-0.18) {PA-IG-DNN};

\draw[black!50,thin]
  (axis cs:89.6,-0.94) -- (axis cs:87.0,-1.55);
\node[ann,anchor=west] at (axis cs:87.0,-1.55) {PA-DeepLIFT-DNN};

\end{axis}
\end{tikzpicture}

\caption{Compression (fidelity trade-offs for all 16
method) model configurations, where rightward movement
indicates greater compression and upward movement indicates
better weighted-F1 preservation. Marker shapes distinguish
conventional, standard-XAI, and PA-XAI methods; open circles
identify the nondominated configurations, and the dashed
line denotes zero weighted-F1 change.}
  \label{fig:module4-reduction}
\end{figure}

Module~4 supplies the utility coordinate $U$: it asks whether a global
attribution ranking retains predictive signal after feature pruning. The
horizontal axis in Figure~\ref{fig:module4-reduction} is the reported percentage
of input dimensions removed; the vertical axis is the median weighted-F1
change, in percentage points, relative to the corresponding full-feature
classifier. Higher and farther right are preferred.

\suppressfloats[t]

PA-SHAP-DNN $(85.3,-0.68)$ and PA-IG-DNN $(90.5,-0.89)$ form the displayed
Pareto set. Relative to PCA $(80.6,-0.69)$, PA-SHAP-DNN adds $4.7$ compression
points with a $+0.01$-point weighted-F1 difference, while PA-IG-DNN adds $9.9$
compression points with a $-0.20$-point difference. At the reported operating
point, the four conventional methods cluster at $80.6\%$ compression with
changes from $-0.87$ to $-0.69$ points. PCA provides a transformed-component
compression benchmark; the attribution methods preserve original-feature
semantics, making the approaches complementary rather than interchangeable.

The remaining PA configurations expose distinct operating regimes. PA-LIME-DNN
$(84.3,-4.27)$ and PA-LIME-RF $(86.0,-2.91)$ select more aggressive compression
points with larger fidelity costs, while PA-DeepLIFT-DNN is dominated by
PA-IG-DNN. These comparisons identify the configurations that extend the
observed frontier and quantify the fidelity cost of additional compression.
Appendix~\ref{app:reduction} gives the complete coordinates and interpretation.
\section{Integrated Operational Analysis and Limitations}
\label{sec:operational-limitations}

The four modules test distinct prerequisites for calibrated operational
reliance. Module~1 evaluates model alignment, repeatability, and prototype cost
($(F,S,T)$); Module~2 measures stability under attack followed by repair ($R$);
Module~3 measures corroboration across model--explainer configurations ($C$);
and Module~4 measures predictive utility after pruning ($U$). The audit vector
is intentionally noncompensatory: reliable use requires adequate evidence
across the dimensions relevant to the intended deployment. High agreement can
coexist with weak model alignment, repeatability with attack sensitivity, and
compression utility with limited semantic fidelity.

The evidence supports technical screening of the evaluated configurations
before investigator use. The timing profile distinguishes lower-cost
reference/gradient methods from perturbation-based offline audits, while the
stress, agreement, and utility modules indicate when additional verification is
needed. Dedicated deployment studies can extend the framework
to workflow-level validation.

\subsection{Limitations}
\label{sec:limitations}
The audit tests encoded-feature controls on four tabular benchmarks; support
consistency and raw-flow reconstruction are separate validation targets. Future
work will measure query-violation rates, including faithfulness replacements,
using Equation~\eqref{eq:validity-leakage} and the diagnostics in
Equations~\eqref{eq:leakage-probability} and~\eqref{eq:leakage-fraction}, and
isolate sampling, reference selection, and repair through controlled ablations.
Adaptive valid-set and explanation-targeting attacks remain untested. Exact
replication requires the unreported splits, seeds, predicate lists, explainer
settings, timing environment, and run-level records for uncertainty estimates.
Analyst trust, decision quality, and workflow benefit require human evaluation.

\section{Conclusion}
\label{sec:conclusion}

This work establishes query validity as a first-class component of
feature-attribution methodology. The valid-query theorem formalizes this
dependence: a procedure confined to the declared rule-valid set gives the same
output for models with equal required oracle information on that set. PA-XAI operationalizes this principle
across LIME, SHAP, Integrated Gradients, and DeepLIFT via
control of neighborhoods, coalitions, references, and paths.

Across four NIDS benchmarks, the paired configurations reveal variability
where standard LIME reports perfect repeatability, change the sign of
faithfulness in five of twelve comparisons, and yield mean agreement of
$0.366$ within PA configurations, compared with $0.217$ within standard
configurations. Under attack followed by protocol repair, PA-DeepLIFT and
PA-IG achieve the strongest curve-level attribution retention among the
evaluated methods. In the feature-reduction audit, PA-SHAP-DNN and PA-IG-DNN
define the compression--fidelity frontier.

Together, these results show that explanation quality in structured domains
depends on the states through which explanations are constructed and
evaluated. A feature ranking is credible only when its generating queries,
references, paths, and replacement states are auditable. PA-XAI provides that
framework for network intrusion detection and establishes a methodological
direction for explainable machine learning wherever domain constraints govern
the meaning of model inputs.

\section*{Acknowledgments}

The authors thank the members of the Rivas.AI Lab (\url{https://lab.rivas.ai}) for their support and valuable feedback throughout the project, and the anonymous reviewers for their useful comments.

\label{page:main-end}

\phantomsection\label{page:references-start}
\bibliographystyle{plainnat}
\bibliography{references}

\clearpage
\appendix
\phantomsection\label{page:appendices-start}
\section{Formal Results and Query-Scope Conditions}
\label{app:formal}

We develop the formal query model, validity predicates, and complete proof of
valid-query invariance. The declared rule set operates in encoded feature
space; empirical-support consistency and raw-event reconstruction provide
separate validation criteria.

\subsection{Encoded states and declared rules}
\label{app:formal-state}

\paragraph{Encoded state space.}
Let $d$ be the encoded feature dimension and let $[d]=\{1,\ldots,d\}$.
Continuous scalar coordinates are indexed by $\mathcal{I}_{c}$. The collection $\mathcal{G}_{d}$ contains
disjoint groups of discrete coordinates, including one-hot categorical groups,
and $\mathcal{I}_{c}\mathbin{\dot\cup}\bigcup_{g\in\mathcal{G}_{d}}g=[d]$.
For $i\in\mathcal{I}_{c}$, $\mathcal{X}_{i}\subseteq\mathbb{R}$ is the declared
coordinate domain. For $g\in\mathcal{G}_{d}$,
$\mathcal{D}_{g}\subseteq\mathbb{R}^{|g|}$ is a finite set of allowed encoded
patterns. Equation~\eqref{eq:encoded-space} defines the resulting product
space $\mathcal{X}$. A classifier is a map
$f:\mathcal{X}\rightarrow\mathbb{R}^{K}$, where $K$ is the number of output
score coordinates and the returned vector contains the scores, logits, or
probabilities used by the evaluated procedure.

\paragraph{Rule predicates.}
Each declared rule is a Boolean predicate $r:\mathcal{X}\rightarrow\{0,1\}$.
The following rule-family display groups the four types of encoded constraint.
Its entries are used in the
\hyperref[tab:rule-examples]{rule-example table} below.
\phantomsection\label{display:rule-families}
\begin{align*}
\mathcal{R}_{\mathrm{dom}} &: \text{coordinate and discrete-group domains},\\
\mathcal{R}_{\mathrm{ord}} &: \text{ordered or bounded cross-feature inequalities},\\
\mathcal{R}_{\mathrm{gate}} &: \text{protocol-conditioned implications},\\
\mathcal{R}_{\mathrm{der}} &: \text{equalities for declared dependent fields}.
\end{align*}
The finite union $\mathcal{R}$ of the
\hyperref[display:rule-families]{four rule families} induces the rule-valid set in
Equation~\eqref{eq:rule-valid-set}. For a group $g$, write
$x_g=(x_i)_{i\in g}$. In the generic forms below, $a$, $b$, and $h_j$ are
scalar functions, $c$ and $h$ are Boolean or scalar conditions as required,
and $S_j\subseteq[d]$ indexes the base coordinates of a declared dependent
field. Exact predicates vary by dataset and backend operation;
Appendix~\ref{app:constraints} catalogs the implemented rule families and their
scope.

\begin{center}
\begin{minipage}{0.98\linewidth}
\small
\phantomsection\label{tab:rule-examples}
\textbf{Illustrative rule examples.} The entries illustrate the notation; exact fields vary across datasets.\par\smallskip
\begin{tabular}{@{}p{0.17\linewidth}p{0.29\linewidth}p{0.45\linewidth}@{}}
\toprule
Family & Generic form & Network-feature example \\
\midrule
Domain & $x_i\in\mathcal{X}_i$ or $x_g\in\mathcal{D}_g$ & Flow durations and packet counts are nonnegative; an encoded protocol group is one of its allowed one-hot patterns. \\
Order & $a(x)\geq b(x)$ & Packet-length maximum $\geq$ mean $\geq$ minimum, with a declared standard deviation no greater than the corresponding range. \\
Gate & $c(x)\Rightarrow h(x)=0$ & If the protocol indicator is not TCP, coordinates representing TCP-only flags are set to zero. \\
Derived & $x_j=h_j(x_{S_j})$ & When a packet-length range field is represented, it is recomputed as maximum minus minimum. \\
\bottomrule
\end{tabular}
\end{minipage}
\end{center}

\subsection{Three validity predicates}
\label{app:formal-validity-levels}

\paragraph{Rule-valid.}
A state $x$ is rule-valid under $\mathcal{R}$ exactly when
$x\in\mathcal{V}_{\mathcal{R}}$. This is the validity property enforced by the implemented rule checks.

\paragraph{Support-consistent.}
Let $\mathcal{D}$ be a named reference sample, $\kappa$ a stated context such as
an encoded protocol value, and
$s_{\mathcal{D}}(\cdot\mid\kappa):\mathcal{X}\rightarrow\mathbb{R}$ an
empirical support or conditional-plausibility score with acceptance threshold
$\tau$. The accepted support set is
\begin{equation}
 \mathcal{S}_{\mathcal{D},\kappa,\tau}
 =\{x\in\mathcal{X}:s_{\mathcal{D}}(x\mid\kappa)\geq\tau\}.
 \label{eq:support-consistent}
\end{equation}
A state is support-consistent only relative to the named quadruple
$(\mathcal{D},\kappa,s_{\mathcal{D}},\tau)$. This predicate is separate from rule
validity unless the support test explicitly includes the declared rules.
Marginal sampling from observed values supplies an empirical contrast;
Equation~\eqref{eq:support-consistent} defines the additional test required for
joint support consistency.

\paragraph{Reconstructable.}
Let $\mathcal{Z}_{\mathrm{valid}}$ be raw packet traces or flow records accepted
by a stated parser, simulator, or reconstruction test, and let
$\Phi:\mathcal{Z}_{\mathrm{valid}}\rightarrow\mathcal{X}$ be the complete
feature-extraction and preprocessing map. The reconstructable set is
\begin{equation}
 \mathcal{X}_{\mathrm{rec}}
 =\{\Phi(z):z\in\mathcal{Z}_{\mathrm{valid}}\}.
 \label{eq:reconstructable}
\end{equation}
Equation~\eqref{eq:reconstructable} defines reconstructability through a raw
event whose feature map yields a state in $\mathcal{X}_{\mathrm{rec}}$.
No strict nesting among the three predicates is assumed for the implemented
sets. A sound and complete rule schema would accept each reconstructable state,
while support consistency and reconstructability impose distinct additional
conditions. The experiments evaluate encoded rule controls; support and
reconstruction tests remain extensions of the audit.

\subsection{Explainer queries, oracle information, and transcripts}
\label{app:formal-transcript}

Let $E$ denote either an attribution procedure or an attribution-evaluation
procedure. Its random state $\omega$ belongs to a probability space
$(\Omega,\mathcal{F},\mathbb{P})$; deterministic procedures use a point-mass
choice of $\omega$. A method-specific query object $q\in\mathcal{Q}_{E}$ has a
finite state footprint $\sigma_E(q)\subseteq\mathcal{X}$. The oracle
$\mathcal{O}^{E}_{f}(q)$ returns exactly the model information requested by
$E$, which need not be a scalar prediction.

Queries may be adaptive. Once $\omega$ is fixed, the next-query and halt
policies are deterministic functions of $x$ and the transcript observed so
far; every model-dependent access, including any model-based reference or
background selection, must appear as an oracle request. Let $N_E\geq1$ be the
finite stopping time induced by these policies, and let $\pi_{E,t}$ denote the
query policy at step $t$:
\begin{align}
q_t &= \pi_{E,t}\!\left(x,\omega,
  \mathcal{O}^{E}_{f}(q_1),\ldots,
  \mathcal{O}^{E}_{f}(q_{t-1})\right),
  \label{eq:adaptive-query}\\
\mathsf{T}_{E}(f,x;\omega)
 &= \bigl((q_t,\mathcal{O}^{E}_{f}(q_t))\bigr)_{t=1}^{N_E},
  \qquad
\mathsf{Q}_{E}(f,x;\omega)
 = \bigcup_{t=1}^{N_E}\sigma_E(q_t).
 \label{eq:appendix-transcript}
\end{align}
Equation~\eqref{eq:adaptive-query} specifies adaptive query selection, and
Equation~\eqref{eq:appendix-transcript} records the resulting transcript.
Once $\omega$ is fixed, let $\Psi_E$ be the deterministic readout map in
Equation~\eqref{eq:procedure-readout} that produces the attribution or score:
\begin{equation}
E(f,x;\omega)=\Psi_E\!\left(x,\omega,
\mathsf{T}_{E}(f,x;\omega)\right).
\label{eq:procedure-readout}
\end{equation}

For IG, let $\gamma:[0,1]\rightarrow\mathcal{X}$ denote the numerical
attribution path and let $\alpha_t\in[0,1]$ denote a quadrature node.
The \hyperref[tab:oracle-procedures]{oracle-interface table} distinguishes
score, gradient, and activation requests within this common abstraction.

\begin{center}
\begin{minipage}{0.98\linewidth}
\small
\phantomsection\label{tab:oracle-procedures}
\textbf{Oracle abstraction by procedure.} The entries state what must be
tracked for the formal result; they do not imply that the procedures estimate
the same quantity.\par\smallskip
\begin{tabular}{@{}p{0.15\linewidth}p{0.37\linewidth}p{0.39\linewidth}@{}}
\toprule
Procedure & State footprint $\sigma_E(q)$ & Oracle information $\mathcal{O}^{E}_{f}(q)$ \\
\midrule
LIME & A generated neighborhood state $z$ & The class score or prediction used to fit the local surrogate at $z$ \\
SHAP & One or more background-completed states for a feature coalition & Model scores, or their backend-specific weighted aggregation, for those completed states \\
IG & A path quadrature point $\gamma(\alpha_t)$ & The target-output gradient $\nabla f(\gamma(\alpha_t))$ required by the numerical line integral \\
DeepLIFT & The explained input and reference states used by the backend & Target/reference activations and the difference multipliers used by the propagation rule \\
Evaluation & An ablated, replaced, or coalition-completed evaluation state & The score used by the stated faithfulness or removal metric \\
\bottomrule
\end{tabular}
\end{minipage}
\end{center}

The SHAP row covers backends whose required information can be represented by
completed states and weighted aggregation; containment is evaluated according
to each backend's actual query interface.

The oracle abstraction is intentionally stronger than equality of scalar model
outputs. In particular, equal scores at isolated states do not by themselves
imply equal gradients, activations, or multipliers. For two classifiers
$f,g:\mathcal{X}\rightarrow\mathbb{R}^{K}$, define explainer-oracle
equivalence on the rule-valid set by
\begin{equation}
f\equiv_{E,\mathcal{V}_{\mathcal{R}}}g
\quad\Longleftrightarrow\quad
\mathcal{O}^{E}_{f}(q)=\mathcal{O}^{E}_{g}(q)
\text{ for every }q\in\mathcal{Q}_{E}
\text{ with }\sigma_E(q)\subseteq\mathcal{V}_{\mathcal{R}}.
\label{eq:oracle-equivalence}
\end{equation}
This condition is the one used by Theorem~\ref{thm:valid-query}. It is
stronger than equality of classifier scores whenever the procedure requests
derivatives, activations, multipliers, or other backend-specific information.

\subsection{Valid-query containment and validity leakage}
\label{app:formal-leakage}

Equation~\eqref{eq:valid-query-contained} requires every queried state in a
realized run to satisfy the declared rules:
\begin{equation}
\mathsf{Q}_{E}(f,x;\omega)\subseteq\mathcal{V}_{\mathcal{R}}.
\label{eq:valid-query-contained}
\end{equation}
Equation~\eqref{eq:leakage-indicator} flags a violating run, and
Equation~\eqref{eq:leakage-probability} averages that event over randomness,
where $\mathbf{1}\{\cdot\}$ is the indicator function:
\begin{align}
L_E(f,x;\omega)
 &= \mathbf{1}\!\left\{
 \mathsf{Q}_{E}(f,x;\omega)\nsubseteq\mathcal{V}_{\mathcal{R}}
 \right\},
 \label{eq:leakage-indicator}\\
\Lambda_E(f,x)
 &= \mathbb{P}_{\omega}\!\left[L_E(f,x;\omega)=1\right].
 \label{eq:leakage-probability}
\end{align}
For a nonempty finite transcript, a query-occurrence leakage fraction can also
be defined as
\begin{equation}
\ell_E(f,x;\omega)=
\frac{\displaystyle\sum_{t=1}^{N_E}
\sum_{z\in\sigma_E(q_t)}\mathbf{1}\{z\notin\mathcal{V}_{\mathcal{R}}\}}
{\displaystyle\sum_{t=1}^{N_E}|\sigma_E(q_t)|}.
\label{eq:leakage-fraction}
\end{equation}
Equation~\eqref{eq:leakage-fraction} counts repeated state occurrences
because they can carry repeated algorithmic weight. A weighted or continuous-path version replaces the finite
average with a normalized query measure. Query leakage identifies exposure to off-rule-valid model information. Output
sensitivity additionally requires that the exposed oracle response influence
the final readout. We call $E$ domain-blind at $(f,x)$ only when changing oracle
information attached solely to an off-$\mathcal{V}_{\mathcal{R}}$ state
footprint can change $E(f,x;\omega)$ for an execution allowed by the procedure.
In this study, $\Lambda_E$ and $\ell_E$ are formal diagnostics; empirical
estimation requires logging complete query transcripts.

\subsection{Valid-query invariance}
\label{app:valid-query-proof}

For each $q\in\mathcal{Q}_E$, let $\mathcal{Y}_E(q)$ be the response
space allowed by the backend interface. Let $\mathfrak{O}_E$ contain maps
$\mathcal{O}$ satisfying $\mathcal{O}(q)\in\mathcal{Y}_E(q)$ for every
request. Each classifier $h$ induces $q\mapsto\mathcal{O}^{E}_{h}(q)$, but a
backend-admissible map in $\mathfrak{O}_E$ need not be induced by any
classifier. Replacing $\mathcal{O}^{E}_{f}$ by $\mathcal{O}$ in the query
policy and readout defines $E(\mathcal{O},x;\omega)$.

\paragraph{Procedure assumptions.}
Theorem~\ref{thm:valid-query} treats $E$ as a finite adaptive oracle procedure.
For fixed $(x,\omega)$, its next-query rule, halt rule, and final readout are
deterministic functions of the transcript prefix. The procedure has no direct
model channel outside $\mathcal{O}^{E}_{f}$, and all model-dependent reference,
background, path, or completion choices are represented in the transcript.
These assumptions permit adaptive queries while ruling out procedures that
inspect model identity or hidden state without recording that access. Any
stochasticity in the model or backend must be included in the coupled random
state or represented in the oracle response law; otherwise the deterministic
statement does not apply.

\paragraph{Influential off-rule-valid response.}
A request $q^{\star}$ is \emph{nontrivially influential} at $(x,\omega)$
when $\sigma_E(q^{\star})\nsubseteq\mathcal{V}_{\mathcal{R}}$ and there
exist $\mathcal{O},\widetilde{\mathcal{O}}\in\mathfrak{O}_E$ such that:
(i) the maps agree on every request other than $q^{\star}$; (ii) the coupled
runs share the transcript prefix through the first issuance of $q^{\star}$;
(iii) the replies at $q^{\star}$ differ; and (iv)
$E(\mathcal{O},x;\omega)\neq E(\widetilde{\mathcal{O}},x;\omega)$.
Later query paths may differ because the procedure can adapt to the changed
reply. This definition isolates one off-rule-valid response while holding the
entire remaining oracle map fixed.

\begin{proof}[Full proof of Theorem~\ref{thm:valid-query}]
Run $E$ against $f$ and $g$ with the same target $x$ and random state $\omega$.
We prove by induction that their transcript prefixes are identical. Before the
first request, both prefixes are empty. Assume the prefixes agree through step
$t-1$. Because the next-query and halt rules depend only on $(x,\omega)$ and
that prefix, both runs make the same halt decision. If they halt, the same
readout rule receives the same prefix and returns the same output. If they
continue, both select the same request $q_t$. The run against $f$ is
valid-query contained, so $\sigma_E(q_t)\subseteq\mathcal{V}_{\mathcal{R}}$.
Oracle equivalence in Equation~\eqref{eq:oracle-equivalence} then gives
$\mathcal{O}^{E}_{f}(q_t)=\mathcal{O}^{E}_{g}(q_t)$, extending prefix equality
through step $t$. The run against $f$ halts after finitely many requests;
therefore the run against $g$ reaches the same halt decision at the same step.
Their complete transcripts and outputs are equal.

For the converse, let $\mathcal{O}$ and
$\widetilde{\mathcal{O}}$ witness influence. They differ only at
$q^{\star}$, whose footprint is not rule-valid, so their restrictions to every
valid-footprint request are identical. Condition (iv) nevertheless gives
unequal outputs. Thus rule-valid oracle information alone does not identify the
procedure output. If classifiers $f$ and $g$ induce the two maps, then
$f\equiv_{E,\mathcal{V}_{\mathcal{R}}}g$ and
$E(f,x;\omega)\neq E(g,x;\omega)$, giving the model-level witness. This last
step requires realizability: an arbitrary assignment of scores, gradients,
activations, or multipliers need not be induced by any classifier in the stated
family.
\end{proof}

\paragraph{Stochastic-query extension.}
Draw the same $\omega\sim\mathbb{P}$ for both runs, assume almost-sure finite
termination, and let the outputs and containment indicators be measurable.
This common-randomness construction couples the output laws. With
$D_{f,g}=\{\omega:E(f,x;\omega)\neq E(g,x;\omega)\}$, the deterministic result
applied in either direction gives
\begin{equation}
D_{f,g}\subseteq
\{L_E(f,x;\omega)=1\}\cap\{L_E(g,x;\omega)=1\}.
\label{eq:stochastic-valid-query-inclusion}
\end{equation}
Equation~\eqref{eq:stochastic-valid-query-inclusion} shows that output
disagreement requires leakage in both runs. Consequently,
\begin{equation}
\mathbb{P}_{\omega}(D_{f,g})
\leq \min\{\Lambda_E(f,x),\Lambda_E(g,x)\}.
\label{eq:stochastic-valid-query-bound}
\end{equation}
If $\mu_f=\operatorname{Law}(E(f,x;\omega))$ and
$\mu_g=\operatorname{Law}(E(g,x;\omega))$, the coupling inequality also yields
\begin{equation}
d_{\mathrm{TV}}(\mu_f,\mu_g)
\leq \mathbb{P}_{\omega}(D_{f,g})
\leq \min\{\Lambda_E(f,x),\Lambda_E(g,x)\}.
\label{eq:stochastic-tv-bound}
\end{equation}
Here, $d_{\mathrm{TV}}$ is total variation distance. Thus zero leakage for
either model gives equal outputs almost surely under this coupling and equal
output laws. Positive leakage removes the invariance guarantee but need not change the
output. Estimating the bounds in
Equations~\eqref{eq:stochastic-valid-query-bound}--\eqref{eq:stochastic-tv-bound}
requires complete query transcripts.

\paragraph{Path-, gradient-, and activation-based procedures.}
For value-query procedures, oracle equivalence can reduce to equality of the
requested class scores at every rule-valid queried state. For a numerical path
method with fixed path $\gamma$, quadrature nodes $\alpha_t$, and weights
$w_t$, a coordinatewise attribution has the form
\begin{equation}
A_{i,k}^{(m)}(f;\gamma)=
\sum_{t=1}^{m}w_t
\frac{\partial f_k}{\partial x_i}\!\left(\gamma(\alpha_t)\right)
\dot{\gamma}_i(\alpha_t).
\label{eq:numerical-path-attribution}
\end{equation}
For the quadrature sum in Equation~\eqref{eq:numerical-path-attribution},
the theorem applies when each node is rule-valid and its gradient oracle agrees. Equal scores there are not enough:
on $\mathcal{V}_{\mathcal{R}}=\{(u,v)\in\mathbb{R}^{2}:v=0\}$, the functions
$f(u,v)=0$ and $g(u,v)=v$ agree in value everywhere on the rule-valid set but
have gradients $(0,0)$ and $(0,1)$. For an exact path attribution with
absolutely continuous $\gamma$, define
\begin{equation}
A_{i,k}(f;\gamma)=
\int_0^1
\frac{\partial f_k}{\partial x_i}\!\left(\gamma(t)\right)
\dot{\gamma}_i(t)\,dt.
\label{eq:exact-path-attribution}
\end{equation}
For Equation~\eqref{eq:exact-path-attribution}, containment must hold along
the path and the required gradients must agree for almost every $t$; equality of the integrands then gives
$A_{i,k}(f;\gamma)=A_{i,k}(g;\gamma)$. A sufficient, stronger condition is
that $f$ and $g$ are continuously differentiable and agree on an open
neighborhood of the path. Equality only on a set with empty
interior, discrete coordinates, or boundary points does not imply equal
gradients. Discrete jumps are covered by the endpoint or difference queries
actually used, not by assuming a derivative across the jump. For DeepLIFT, the
oracle must likewise include the backend's reference activations and
multipliers; equality of final scores need not fix those internal quantities.

\subsection{Evaluation metrics as composite query procedures}
\label{app:evaluation-corollary}

Let $M$ first receive the attribution $a=E(f,x;\omega)$ and then issue a finite,
possibly adaptive sequence of metric requests $q^M_1,\ldots,q^M_m$. Here
$\omega$ contains all explainer- and metric-side randomness. Write
\begin{equation}
M(E,f,x;\omega)=\Psi_M\!\left(
 x,\omega,a,(q^M_j,\mathcal{O}^{M}_{f}(q^M_j))_{j=1}^{m}\right).
\label{eq:appendix-evaluation-metric}
\end{equation}
In Equation~\eqref{eq:appendix-evaluation-metric}, a request can encode a
feature-removal state $r_j(x,a)$, a replacement state,
a conditional completion, or another model-dependent evaluation query. Form a
composite procedure $H_{E,M}$ that records the complete transcript of the
explainer followed by the metric, including model-dependent choices made after
observing $a$ or an earlier metric reply.

\begin{proof}[Full proof of Corollary~\ref{cor:evaluation-query}]
Fix $(x,\omega)$. The composite $H_{E,M}$ is a finite adaptive oracle
procedure of the form used by Theorem~\ref{thm:valid-query}. If its complete
queried-state set is contained in $\mathcal{V}_{\mathcal{R}}$ and the two models
return equal explainer- and metric-specific oracle information on every
rule-valid request, Theorem~\ref{thm:valid-query} gives identical composite
transcripts. The attribution, metric requests, replies, and readout are
therefore identical, so the metric values are equal. Conversely, if a
backend-admissible off-rule-valid metric request is nontrivially influential,
change only that reply while holding every rule-valid request fixed. The metric
changes although the valid-request restrictions agree. A classifier-level
witness again requires classifiers that induce both oracle maps.
\end{proof}

The reported faithfulness protocol replaces selected coordinates with baseline
mean values. If $J\subseteq[d]$ is the selected coordinate set and $\mu$ is the
baseline vector, the replacement state has the generic form
\begin{equation}
[r_{J,\mu}(x)]_i=
\begin{cases}
\mu_i, & i\in J,\\
x_i, & i\notin J.
\end{cases}
\label{eq:mean-replacement-state}
\end{equation}
The mean replacement in Equation~\eqref{eq:mean-replacement-state} can
violate cross-feature relations. The corollary
therefore makes replacement-state validity part of the metric specification.
Recording complete replacement transcripts and rule-check outcomes would permit
direct estimation of evaluation-query leakage.

\subsection{Attack followed by repair versus adaptive valid-set optimization}
\label{app:formal-attack}

Let $A_{\varepsilon}(f,x,y)$ denote the reported classifier-evasion routine and
let $\Pi_{\mathcal R}$ denote the implemented ordered repair map. The evaluated
composition is
\begin{equation}
 x_{\mathrm{atk}}=A_{\varepsilon}(f,x,y),\qquad
 x_{\mathrm{rep}}=\Pi_{\mathcal R}(x_{\mathrm{atk}}).
\label{eq:formal-attack-repair}
\end{equation}
The notation $\Pi_{\mathcal R}$ denotes the implemented rule-repair procedure,
which is distinct from an orthogonal projection or adversarial optimizer.
Repair-aware distance control can be added explicitly when preservation of
$\rho(x_{\mathrm{rep}},x)\leq\varepsilon$ is required.

An adaptive valid-set classifier-evasion attack would instead solve, or
approximate during optimization,
\begin{equation}
 x^{\star}_{\mathrm{ad}}\in
 \arg\max_{\substack{z\in\mathcal V_{\mathcal R}\\
 \rho(z,x)\leq\varepsilon}}
 \mathcal L_{\mathrm{cls}}(f(z),y).
\label{eq:formal-adaptive-valid-set}
\end{equation}
Equation~\eqref{eq:formal-attack-repair} characterizes the evaluated repair
pipeline, whereas Equation~\eqref{eq:formal-adaptive-valid-set} characterizes
its adaptive valid-set extension. Attribution-similarity, top-$k$ replacement,
rank-change, and consensus-disruption losses provide additional
explanation-targeting objectives.

\subsection{Audit evidence profile}
\label{app:audit-profile}

For a fixed evaluation record set $\mathcal{D}$, the audit profile in
Equation~\eqref{eq:audit-profile} is an ordered diagnostic record. Its
components may themselves be vectors or tables:
\begin{equation}
\mathbf{A}(E,f,\mathcal{D})
\in
\mathcal{Y}_{F}\times\mathcal{Y}_{S}\times\mathcal{Y}_{R}
\times\mathcal{Y}_{C}\times\mathcal{Y}_{U}\times\mathcal{Y}_{T}.
\label{eq:audit-profile-product}
\end{equation}
Here $F$ stores model-alignment evidence under the stated faithfulness
protocol; $S$ stores repeated-explanation stability; $R$ stores attribution
similarity and perturbed-feature ranking under attack followed by protocol
repair; $C$ stores cross-method agreement and interaction comparisons; $U$ stores feature-reduction utility; and $T$ stores measured
computational cost. The module map is
\begin{equation}
\mathrm{M1}\mapsto(F,S,T),\qquad
\mathrm{M2}\mapsto R,\qquad
\mathrm{M3}\mapsto C,\qquad
\mathrm{M4}\mapsto U.
\label{eq:audit-module-map}
\end{equation}
The spaces in Equation~\eqref{eq:audit-profile-product} have different units
and answer different scientific questions; Equation~\eqref{eq:audit-module-map}
assigns their module roles. The product representation preserves the evidence types
without forcing a scalarization; operational reliance is assessed against the
subset of coordinates required by the intended use.

\paragraph{Scientific scope.}
The theorem and corollary establish transcript invariance under explicit
query-containment and oracle-equivalence conditions. Their role in this paper is
to connect a structured-network rule set with one oracle abstraction spanning
neighborhood, coalition, path/reference, propagation, and evaluation queries.
The four empirical modules then test how those choices alter model alignment,
stress behavior, corroboration, and downstream utility.
\section{PA-XAI Algorithms and Backend Scope}
\label{app:methods}

This appendix specifies the procedure-level logic for each PA-XAI instance.
LIME, SHAP, Integrated Gradients (IG), and DeepLIFT refer to their established
explainer families
\cite{ribeiro2016should,lundberg2017unified,sundararajan2017axiomatic,shrikumar2017learning};
\emph{standard} and \emph{PA} identify the paired configurations evaluated in
this study.

\subsection{Common interface and ordered repair}
\label{app:methods-common}
Let $\mathcal D_{\mathrm{tr}}$ be the training set, $x$ the encoded record to
explain, and $F=f_k$ the scalar model output being attributed. Let $P$ denote
the encoded protocol field or group, $C$ the continuous coordinates, and
$G_1,\ldots,G_m$ the discrete coordinates or groups. The ordered repair map
$\Pi_{\mathcal R}$ implements the declared rule set through domain handling,
restoration of ordered statistics, protocol-gated zeroing, and recomputation of
derived fields. Its symbol denotes a deterministic repair routine rather than a
nearest-point or optimization operator.

\subsection{PA-LIME: conditioned marginals, repair, and surrogate fit}
\label{app:pa-lime}

\paragraph{Algorithm B.1 (PA-LIME).}
\textbf{Input:} $x$, output $F$, $\mathcal D_{\mathrm{tr}}$, neighborhood size
$N$, the continuous perturbation rule, the LIME kernel $K_x$, and
$\Pi_{\mathcal R}$. \textbf{Output:} the fitted local surrogate $\widehat g$ and
its coefficients.
\begin{enumerate}
  \item Draw a protocol value
  $p_j\sim\widehat P_{\mathcal D_{\mathrm{tr}}}(P)$ and form
  $\mathcal D_{p_j}=\{z\in\mathcal D_{\mathrm{tr}}:P(z)=p_j\}$.
  \item Draw each discrete coordinate or group from its empirical marginal in
  $\mathcal D_{p_j}$ and perturb the continuous coordinates with the study's
  Gaussian rule, producing $z_j^{\mathrm{raw}}$.
  \item Apply the ordered repair once,
  $z_j=\Pi_{\mathcal R}(z_j^{\mathrm{raw}})$, and evaluate $F(z_j)$.
  \item Fit the same weighted surrogate used by the paired LIME comparator:
  \begin{equation}
    \widehat g\in\arg\min_{g\in\mathcal G}
    \sum_{j=1}^{N}K_x(z_j)
    \bigl[F(z_j)-g(\xi(z_j))\bigr]^2+\Omega(g),
    \label{eq:pa-lime-objective}
  \end{equation}
  where $\xi$ and $\mathcal G$ are the comparator's interpretable encoding and
  surrogate family.
\end{enumerate}

\paragraph{Controlled property and scope.}
Equation~\eqref{eq:pa-lime-objective} fits the surrogate to scores at repaired
neighbors. Rows accepted by the checker satisfy the implemented constraints. The sampler
uses protocol-conditioned marginals rather than a joint conditional generator,
so empirical-support and reconstruction tests remain separate validation
stages. The sampled protocol follows the training distribution and may differ
from $P(x)$; Table~\ref{tab:method-configuration} consolidates the remaining
configuration parameters.

\subsection{PA-SHAP: benign references and backend-specific completion}
\label{app:pa-shap}
Let
\begin{equation}
\mathcal B_{\mathrm{ben}}
=\{b\in\mathcal D_{\mathrm{tr}}:b\text{ is assigned to the benign class}\}
\label{eq:benign-background}
\end{equation}
be the benign training background. Equation~\eqref{eq:benign-background}
sets the reference population used in coalition completion. It spans the protocols represented in the
training data and therefore defines a contrast against observed benign records.

\paragraph{Algorithm B.2 (KernelSHAP branch).}
For coalition mask $m$, target $x$, and $b\in\mathcal B_{\mathrm{ben}}$, form
\begin{equation}
 z^{\mathrm{raw}}(m;x,b)=m\odot x+(1-m)\odot b,
 \qquad
 z_{\mathcal R}(m;x,b)=\Pi_{\mathcal R}
 \bigl(z^{\mathrm{raw}}(m;x,b)\bigr).
 \label{eq:pa-kernel-completion}
\end{equation}
The KernelSHAP regression receives the repaired-completion game
\begin{equation}
 v_{\mathcal R}(m)
 =\mathbb E_{b\sim\mathcal B_{\mathrm{ben}}}
 \left[F\bigl(z_{\mathcal R}(m;x,b)\bigr)\right].
 \label{eq:pa-kernel-game}
\end{equation}
Standard KernelSHAP weighting and regression are then applied to these values.

\paragraph{TreeSHAP and DeepSHAP branches.}
TreeSHAP and DeepSHAP use the benign background through their native backend
semantics. KernelSHAP receives explicit coalition repair; the tree and gradient
backends therefore control the reference distribution rather than every
internal query state.

\paragraph{Controlled property and scope.}
A repaired KernelSHAP completion satisfies the implemented checks. Because
repair can update multiple coordinates to restore cross-feature relations,
Equations~\eqref{eq:pa-kernel-completion}--\eqref{eq:pa-kernel-game} define a
repaired-completion game. Table~\ref{tab:method-configuration} records the
backend, sampling, and grouping parameters needed to reproduce that game.

\subsection{Decision-boundary reference shared by PA-IG and PA-DeepLIFT}
\label{app:boundary-reference}
Let $s_c(z)$ be the logit for class $c$, $\tau_c$ its decision threshold, and
$\mathcal D_{\mathrm{tr},c}$ the class-specific candidate subset. The reference
selection rule is
\begin{equation}
 b_c\in\arg\min_{z\in\mathcal D_{\mathrm{tr},c}}
 \left|s_c(z)-\tau_c\right|.
 \label{eq:boundary-reference}
\end{equation}
Thus, the PA reference is an observed training row near the decision boundary,
whereas the paired standard configurations use the zero vector. A zero
reference is a study choice rather than a defining requirement of IG or
DeepLIFT. Table~\ref{tab:method-configuration} records the candidate-class,
tie, protocol-matching, and output-specific reference choices.

\subsection{PA-IG: continuous segment and discrete endpoint changes}
\label{app:pa-ig}
\paragraph{Algorithm B.3 (PA-IG).}
Fix $b=b_c$ from Equation~\eqref{eq:boundary-reference}. Let $D$ denote the
union of discrete groups and define $z^{(0)}=(x_C,b_D)$. The continuous path is
\begin{equation}
 \gamma_C(t)_i=
 \begin{cases}
 b_i+t(x_i-b_i), & i\in C,\\
 b_i, & i\in D,
 \end{cases}
 \qquad t\in[0,1].
 \label{eq:pa-ig-continuous-path}
\end{equation}
For $i\in C$, the continuous contribution is
\begin{equation}
 \phi_i^{C}=(x_i-b_i)
 \int_0^1\frac{\partial F(\gamma_C(t))}{\partial x_i}\,dt.
 \label{eq:pa-ig-continuous-attribution}
\end{equation}
Choose a fixed order $\pi$ over the discrete groups. From $z^{(0)}$, construct
$z^{(j)}$ by replacing group $G_{\pi_j}$ with its target value, leaving
not-yet-changed groups at the reference value. The group contribution is
\begin{equation}
 \Phi_{G_{\pi_j}}^{D}=F(z^{(j)})-F(z^{(j-1)}),
 \qquad j=1,\ldots,m.
 \label{eq:pa-ig-discrete-attribution}
\end{equation}
The method yields a group-level discrete contribution; allocation across the
encoded columns of a group is left unspecified.

\paragraph{Conditional summation identity.}
If $F\circ\gamma_C$ is absolutely continuous, the integral is exact, all
endpoint differences use the same output $F$, and $z^{(m)}=x$, then
\begin{align}
 \sum_{i\in C}\phi_i^C
 &=F(z^{(0)})-F(b),\label{eq:pa-ig-continuous-sum}\\
 \sum_{j=1}^{m}\Phi_{G_{\pi_j}}^D
 &=F(x)-F(z^{(0)}),\label{eq:pa-ig-discrete-sum}\\
 \sum_{i\in C}\phi_i^C+
 \sum_{j=1}^{m}\Phi_{G_{\pi_j}}^D
 &=F(x)-F(b).
 \label{eq:pa-ig-summation}
\end{align}
Equations~\eqref{eq:pa-ig-continuous-path} and~\eqref{eq:pa-ig-continuous-attribution}
define the continuous segment and its contributions;
Equation~\eqref{eq:pa-ig-discrete-attribution} assigns the endpoint changes.
Equations~\eqref{eq:pa-ig-continuous-sum} and~\eqref{eq:pa-ig-discrete-sum}
telescope those parts, yielding Equation~\eqref{eq:pa-ig-summation}.
A numerical residual analysis would quantify finite-quadrature error.

\paragraph{Controlled property and scope.}
The continuous segment keeps discrete groups at valid endpoint values and the
discrete contribution uses endpoint changes rather than fractional
interpolation. Transition order can redistribute contribution among interacting
discrete groups while preserving the total in
Equation~\eqref{eq:pa-ig-summation}. Protocol matching, shortest-path
construction, explicit path-state checking, empirical-support constraints, and
reference optimization are complementary design choices.

\subsection{PA-DeepLIFT: decision-boundary reference and endpoint multipliers}
\label{app:pa-deeplift}

\paragraph{Algorithm B.4 (PA-DeepLIFT).}
\textbf{Input:} $x$, output $F$, and $b=b_c$ from
Equation~\eqref{eq:boundary-reference}. \textbf{Output:} DeepLIFT Rescale
contributions relative to $b$.
\begin{enumerate}
  \item Evaluate the network at $x$ and $b$ and form activation differences
  relative to the reference pass.
  \item Apply the DeepLIFT Rescale rules to obtain input-to-output multipliers
  $m_i^{F}(x,b)$.
  \item Return
  \begin{equation}
    \phi_i^{\mathrm{DL}}=m_i^{F}(x,b)(x_i-b_i).
    \label{eq:pa-deeplift}
  \end{equation}
  Discrete coordinates enter through endpoint differences, so no interpolated
  discrete input is scored.
\end{enumerate}

\paragraph{Controlled property and scope.}
Equation~\eqref{eq:pa-deeplift} applies the Rescale multiplier to each
endpoint difference. The PA change is reference selection in
Equation~\eqref{eq:boundary-reference}; the propagation rule remains unchanged. The observed reference removes the zero-vector contrast and preserves
endpoint semantics. Protocol matching, explicit reference validation, and
reference optimization are configurable extensions. The Rescale backend
preserves its standard summation behavior for supported operations; a numerical
residual analysis would quantify implementation-level approximation.

\subsection{Implementation logic and configuration parameters}
\label{app:methods-configuration}
Table~\ref{tab:method-configuration} separates the specified algorithmic logic
from parameters required for exact replication.

\begin{table}[h]
\centering
\small
\caption{Method configuration and reproducibility details. Unspecified fields
are identified directly rather than inferred.}
\label{tab:method-configuration}
\begin{tabular}{@{}p{0.18\linewidth}p{0.36\linewidth}p{0.38\linewidth}@{}}
\toprule
Procedure & Specified algorithmic logic & Additional replication detail \\
\midrule
PA-LIME & Protocol draw; protocol-filtered discrete marginals; Gaussian continuous perturbation; ordered repair; weighted surrogate & Neighborhood count; Gaussian scales; kernel and sparsity settings; dataset-specific rule map \\
Kernel PA-SHAP & Benign background; coalition completion; ordered repair before scoring & Background count; coalition count; mask grouping; repaired-coordinate policy \\
Tree/Deep PA-SHAP & Benign background passed to the native backend & Backend marginalization mode and internal query-state diagnostics \\
PA-IG & Threshold-near class reference; continuous segment; ordered discrete endpoint differences; conditional summation identity & Candidate-class convention; ties; protocol matching; discrete order and group-to-column allocation; quadrature rule \\
PA-DeepLIFT & Threshold-near reference; Rescale endpoint differences & Candidate-class convention; ties; protocol matching; backend/operator coverage and residual analysis \\
\bottomrule
\end{tabular}
\end{table}
\section{Declared Constraint Catalog and Validity Scope}
\label{app:constraints}

PA-XAI applies four families of encoded-space constraints: domain, order,
protocol-gating, and derived-feature rules. Their exact feature instantiations
vary by dataset and backend operation.

\subsection{Constraint families}
\label{app:constraints-families}
\begin{itemize}
  \item \textbf{Domain rules:} nonnegativity and finite allowed patterns for
  encoded discrete groups.
  \item \textbf{Order rules:} relations such as maximum $\geq$ mean $\geq$
  minimum and a stated standard deviation bounded by the corresponding range.
  \item \textbf{Protocol gates:} protocol-conditioned zeroing of fields that
  are inactive under the selected encoded protocol.
  \item \textbf{Derived-field rules:} recomputation of declared arithmetic
  fields, such as a range from maximum minus minimum.
\end{itemize}
The generic forms and illustrative examples appear in
Appendix~\ref{app:formal-state}.

\subsection{Repair order and operation-specific scope}
\label{app:constraints-repair}
The ordered repair routine applies domain handling, restores ordered statistics,
clamps the reported standard deviation to its range, enforces protocol-gated
zeros, and recomputes derived fields before model evaluation. Explicit
state-level repair is applied to PA-LIME neighborhoods and KernelSHAP coalition
completions. TreeSHAP and DeepSHAP control the background distribution through
their native backend semantics. PA-IG and PA-DeepLIFT control the reference or
path construction rather than every internal query state.

\subsection{Dataset instantiation}
\label{app:constraints-datasets}
The four constraint families are instantiated on NSL-KDD, CIC-IDS-2017,
UNSW-NB15, and CSE-CIC-IDS2018. Exact encoded feature maps and
dataset-specific predicate lists are not reported, so the catalog specifies
rule families rather than a reproducible predicate map for each dataset.
The positive guarantee is
membership in the declared encoded-space rule set for operations that pass
through the checker.

\subsection{Validation layers}
\label{app:constraints-validity}
Rule validity, empirical-support consistency, and raw-event reconstructability
are separate predicates, as formalized in
Appendix~\ref{app:formal-validity-levels}. The first verifies explicit
encoded-feature relations; the second evaluates plausibility under a named
support model; the third links an encoded state to a valid raw event through a
complete feature map. These complementary checks have separate acceptance
criteria; the present experiments do not measure support consistency or
raw-event reconstructability.
\section{Experimental Configuration and Reproducibility}
\label{app:setup}

This appendix consolidates the experimental configuration used across the four
audit modules. Fields listed as additional replication details were not
reported; common benchmark splits or software defaults are not assumed.

\subsection{Data scope and preprocessing}
\label{app:setup-data}
The study uses NSL-KDD, CIC-IDS-2017, UNSW-NB15, and CSE-CIC-IDS2018 as tabular
benchmark corpora. Median imputation, one-hot encoding, and min--max scaling
precede classification. Training partitions define PA-LIME distributions,
PA-SHAP backgrounds, and PA-IG/PA-DeepLIFT references; Module~4 uses a
calibration set and a held-out test partition.

Table~\ref{tab:setup-datasets} maps each benchmark to its role in the
four-module evaluation and identifies the dataset-specific information needed
for exact replication.

\begin{table}[h]
\centering
\caption{Dataset roles and configuration details.}
\label{tab:setup-datasets}
\footnotesize
\setlength{\tabcolsep}{4pt}
\begin{tabular}{@{}p{0.23\linewidth}p{0.26\linewidth}p{0.43\linewidth}@{}}
\toprule
\textbf{Dataset} & \textbf{Role} & \textbf{Additional detail for exact replication} \\
\midrule
NSL-KDD & Four-module benchmark & Release identifier, class mapping, split indices, encoded dimensions, and preprocessing-fit scope \\
CIC-IDS-2017 & Four-module benchmark & Release identifier, class mapping, split indices, encoded dimensions, and preprocessing-fit scope \\
UNSW-NB15 & Four-module benchmark & Release identifier, class mapping, split indices, encoded dimensions, and preprocessing-fit scope \\
CSE-CIC-IDS2018 & Four-module benchmark; domain-reference-set analysis & Release identifier, class mapping, split indices, encoded dimensions, preprocessing-fit scope, and reference-set definitions \\
\bottomrule
\end{tabular}
\end{table}

\subsection{Classifiers and explainer coverage}
\label{app:setup-models}
The primary classifier families are a fully connected DNN, RF, and XGB. The
DNN is also the attacked classifier in Module~2. Module~4 uses a separate RF
proxy for pruning and retrains a fully connected DNN, 1D-CNN, RF, and SVM on
the selected features.

Table~\ref{tab:setup-coverage} distinguishes the primary
classifier--explainer pairings from the separate attack, pruning, and
downstream-model roles.

\begin{table}[h]
\centering
\caption{Model--explainer coverage and downstream roles. ``Std/PA'' means that
both paired configurations appear in the explanation-quality results.}
\label{tab:setup-coverage}
\footnotesize
\setlength{\tabcolsep}{5pt}
\begin{tabular}{@{}p{0.22\linewidth}p{0.40\linewidth}p{0.27\linewidth}@{}}
\toprule
\textbf{Classifier family} & \textbf{Explanation-quality methods} &
\textbf{Additional role} \\
\midrule
DNN & SHAP, LIME, IG, DeepLIFT (Std/PA) & Attacked classifier; downstream feature-reduction classifier \\
Random Forest & SHAP, LIME (Std/PA) & Distinct pruning proxy; downstream classifier \\
XGBoost & SHAP, LIME (Std/PA) & Primary explanation-quality classifier \\
1D-CNN & -- & Downstream feature-reduction classifier \\
SVM & -- & Downstream feature-reduction classifier \\
\bottomrule
\end{tabular}
\end{table}
Table~\ref{tab:repro-status} consolidates the remaining training,
checkpoint, threshold, and backend details needed for exact replication.

\subsection{Module 1: explanation-quality protocol}
\label{app:setup-quality}
\begin{itemize}
  \item \textbf{Faithfulness:} replace top-ranked features with baseline means,
  measure the true-class score decrease, and compute Pearson correlation with
  mean absolute attribution.
  \item \textbf{Stability:} compute Jaccard overlap between top-$k$ feature sets
  across repeated explanations.
  \item \textbf{Complexity:} normalized Shannon entropy of the attribution
  vector.
  \item \textbf{Sparsity:} fraction of attribution magnitudes below a threshold.
  \item \textbf{Efficiency:} per-sample prototype latency.
\end{itemize}
Table~\ref{tab:repro-status} consolidates the remaining $k$, replacement,
sampling, seed, sparsity, and timing details. Cross-method completeness is
omitted because the four explainer families use different reported semantics.

\subsection{Module 2: classifier evasion followed by repair}
\label{app:setup-attacks}
The attacked model is the DNN under full parameter and gradient access. FGSM is
evaluated at $\varepsilon\in\{0.01,0.05,0.1,0.2,0.3\}$ and PGD at
$\varepsilon\in\{0.01,0.05,0.1\}$. The unrestricted regime permits changes to
all encoded features; the repair regime passes the same attack output through
$\Pi_{\mathcal R}$. Evasion success is the fraction of initially correct
samples that become misclassified. Attribution comparison uses cosine
similarity, a mean Lipschitz ratio, and class-equivalent top-$k$ Jaccard;
perturbed-feature localization uses Top-1, Top-3, and mean attribution rank.
Table~\ref{tab:repro-status} consolidates the remaining optimizer, sample,
weighting, and post-repair distance details.

\subsection{Module 3: agreement, interactions, and domain references}
\label{app:setup-agreement}
Pairwise agreement uses $10{,}000$ samples per dataset and 66 configuration
pairs. The metrics are Spearman $\rho$, Kendall $\tau_b$, top-5 and top-10
overlap, and a Wilcoxon signed-rank test at $\alpha=0.05$. Interaction values
use $5{,}000$ samples per dataset. On CSE-CIC-IDS2018, RRA@15 measures recovery
of predefined 6--8-feature domain sets among the top 15 attributions.
Table~\ref{tab:repro-status} consolidates the configuration, sampling,
multiple-comparison, interaction, and domain-set details.

\subsection{Module 4: attribution-guided feature reduction}
\label{app:setup-reduction}
The procedure has three stages:
\begin{enumerate}
  \item Compute a global ranking from mean absolute attribution over $5{,}000$
  calibration samples.
  \item Prune features with a coarse-to-fine schedule using an RF proxy with 50
  estimators and maximum depth 10, trained on a $50{,}000$-sample subsample.
  Stop when macro-F1 decreases by more than 0.01 from the full-feature proxy
  baseline.
  \item Retrain a fully connected DNN, 1D-CNN, RF, and SVM on the selected
  subset and assess performance on a held-out test partition. Apply the same
  procedure to chi-squared, Information Gain, PCA, and Spearman baselines.
\end{enumerate}
Table~\ref{tab:repro-status} consolidates the pruning, calibration, proxy,
downstream-model, test-index, selected-feature, and matching details.

\subsection{Software, hardware, and execution environment}
\label{app:setup-runtime}
The prototype reports per-sample latency. Table~\ref{tab:repro-status}
identifies the unreported execution environment, timing protocol, software,
schemas, checkpoints, split indices, configurations, and run-level outputs.
The latency values describe this prototype; cross-system comparisons require
the hardware, software versions, and timing harness.

\subsection{Reproducibility summary}
\label{app:reproducibility}
\begin{table}[h]
\centering
\caption{Reported configuration and additional detail required for exact
replication.}
\label{tab:repro-status}
\footnotesize
\setlength{\tabcolsep}{4pt}
\begin{tabular}{@{}p{0.24\linewidth}p{0.31\linewidth}p{0.37\linewidth}@{}}
\toprule
\textbf{Component} & \textbf{Reported configuration} & \textbf{Additional replication detail} \\
\midrule
Data preparation & Four datasets; median imputation; one-hot encoding; min--max scaling & Release IDs, split indices and seeds, label maps, encoded schemas, transform-fit scope \\
Models and explainers & Model/explainer families and PA procedure logic & Architectures, hyperparameters, seeds, checkpoints, backend map \\
Module sampling & Agreement/interactions and reduction calibration/proxy counts & Module-1/2 sample indices and repeat schedules \\
Attacks and reduction & Budgets, attack-then-repair order, proxy RF, stopping rule, downstream families, baselines & Optimizer details, repair-distance audit, exact matched subsets \\
Systems & Prototype latency & Hardware, software versions, timing harness, code, environment, logs \\
\bottomrule
\end{tabular}
\end{table}
\section{Full Explanation-Quality Results}
\label{app:quality}

\paragraph{Metric scope.}
Module~1 supplies the audit coordinates $(F,S,T)$ in
Equation~\eqref{eq:audit-profile}. Faithfulness replaces top-ranked features
with baseline means and computes the Pearson correlation between attribution
magnitude and the resulting score change. Corollary~\ref{cor:evaluation-query}
places these replacement states inside the same query-level analysis as the
explainer. Faithfulness therefore evaluates both the ranking and the specified
replacement procedure; logging rule-check outcomes would support a direct
replacement-validity analysis.

Repeated-run stability is the Jaccard overlap of top-$k$ feature sets across
repeated explanations. For deterministic SHAP, IG, and DeepLIFT computations,
$J=1.0$ records exact repeatability under the fixed configuration. Exact $k$,
repeat count, and seed schedule are additional replication details.

Table~\ref{tab:quality-averages} summarizes the four-dataset mean
faithfulness and repeatability values for every retained standard/PA pairing.
The changes are method-specific, ranging from a $-0.35$ faithfulness shift for
DNN-LIME to a $+0.13$ shift for XGB-SHAP.

\begin{table}[t]
\centering
\caption{Four-dataset mean explanation-quality results. Faithfulness is the
reported Pearson mean-replacement correlation $r$; $J$ is repeated-run top-$k$
Jaccard. Deltas are PA minus standard and are recomputed from displayed
endpoints.}
\label{tab:quality-averages}
\small
\setlength{\tabcolsep}{3.2pt}
\begin{tabular}{@{}llrrrrrr@{}}
\toprule
& & \multicolumn{3}{c}{Faithfulness $r$} & \multicolumn{3}{c}{Repeatability $J$} \\
\cmidrule(lr){3-5}\cmidrule(lr){6-8}
Model & Method & Std & PA & $\Delta$ & Std & PA & $\Delta$ \\
\midrule
DNN & SHAP & 0.63 & 0.62 & -0.01 & 1.00 & 1.00 & 0.00 \\
DNN & LIME & 0.28 & -0.07 & -0.35 & 0.63 & 0.46 & -0.17 \\
DNN & IG & 0.57 & 0.43 & -0.14 & 1.00 & 1.00 & 0.00 \\
DNN & DeepLIFT & 0.51 & 0.54 & +0.03 & 1.00 & 1.00 & 0.00 \\
\midrule
RF & SHAP & 0.79 & 0.80 & +0.01 & 1.00 & 1.00 & 0.00 \\
RF & LIME & 0.38 & 0.49 & +0.11 & 0.69 & 0.74 & +0.05 \\
\midrule
XGB & SHAP & 0.58 & 0.71 & +0.13 & 1.00 & 1.00 & 0.00 \\
XGB & LIME & 0.23 & 0.18 & -0.05 & 0.65 & 0.54 & -0.11 \\
\bottomrule
\end{tabular}
\end{table}

\begin{table}[t]
\centering
\caption{Mean prototype time per explanation in milliseconds. Exact hardware,
software, repetition count, and timing protocol are additional replication
details.}
\label{tab:quality-latency}
\small
\setlength{\tabcolsep}{8pt}
\begin{tabular}{@{}llrr@{}}
\toprule
Model & Method & Standard & PA \\
\midrule
DNN & SHAP & 65.10 & 828.8 \\
DNN & LIME & 92.20 & 1690.6 \\
DNN & IG & 0.70 & 54.6 \\
DNN & DeepLIFT & 0.01 & 6.7 \\
\midrule
RF & SHAP & 13.90 & 248.5 \\
RF & LIME & 127.30 & 1738.0 \\
\midrule
XGB & SHAP & 0.10 & 1990.7 \\
XGB & LIME & 123.40 & 1737.4 \\
\bottomrule
\end{tabular}
\end{table}

\begin{table}[t]
\centering
\caption{Per-dataset LIME faithfulness and repeated-run Jaccard under the
standard and PA neighborhood procedures. CSE denotes CSE-CIC-IDS2018.}
\label{tab:lime-per-dataset}
\scriptsize
\setlength{\tabcolsep}{2.4pt}
\begin{tabular}{@{}lrrrrrrrr@{}}
\toprule
& \multicolumn{2}{c}{NSL-KDD} & \multicolumn{2}{c}{CIC-IDS-2017} & \multicolumn{2}{c}{UNSW-NB15} & \multicolumn{2}{c}{CSE} \\
\cmidrule(lr){2-3}\cmidrule(lr){4-5}\cmidrule(lr){6-7}\cmidrule(lr){8-9}
& Std & PA & Std & PA & Std & PA & Std & PA \\
\midrule
\multicolumn{9}{l}{\emph{Faithfulness $r$}} \\
DNN/LIME & -0.05 & 0.20 & 0.50 & -0.05 & 0.19 & -0.14 & 0.49 & -0.30 \\
RF/LIME & 0.18 & 0.66 & 0.31 & 0.39 & 0.66 & 0.56 & 0.36 & 0.36 \\
XGB/LIME & 0.14 & 0.38 & 0.21 & 0.29 & 0.25 & 0.10 & 0.31 & -0.07 \\
\midrule
\multicolumn{9}{l}{\emph{Repeatability $J$}} \\
DNN/LIME & 0.17 & 0.52 & 0.37 & 0.45 & 1.00 & 0.48 & 1.00 & 0.39 \\
RF/LIME & 0.23 & 0.74 & 0.52 & 0.67 & 1.00 & 0.69 & 1.00 & 0.86 \\
XGB/LIME & 0.15 & 0.62 & 0.44 & 0.50 & 1.00 & 0.43 & 1.00 & 0.60 \\
\bottomrule
\end{tabular}
\end{table}

\paragraph{LIME stability.}
Table~\ref{tab:lime-per-dataset} contains the six standard-LIME values equal to
$1.0$: all three classifier families on UNSW-NB15 and CSE-CIC-IDS2018. Their
paired PA-LIME values average $0.575$, for a difference of $-0.425$. The
aggregate results establish sensitivity to neighborhood construction; retaining
the generated neighborhoods, rule-pass indicators, local-surrogate fit errors,
and model-output spread would enable a mechanistic decomposition.

\paragraph{Faithfulness.}
Five of the 12 LIME model--dataset comparisons change sign, including every DNN
comparison. The DNN average ranking changes from SHAP ($0.63$), IG ($0.57$),
DeepLIFT ($0.51$), and LIME ($0.28$) to SHAP ($0.62$), DeepLIFT ($0.54$), IG
($0.43$), and LIME ($-0.07$). These rankings characterize model-response
alignment under the reported mean-replacement protocol. Negative coefficients
indicate inverse association between attribution magnitude and score change.

\paragraph{Latency and secondary metrics.}
Table~\ref{tab:quality-latency} reports four-dataset mean prototype times. The
values distinguish lower-latency gradient/reference methods from the
perturbation-based configurations. Hardware, software, repetition count, and
timing variance are listed in Appendix~\ref{app:setup-runtime} as additional
replication details. The study also evaluates normalized-entropy complexity and sparsity as
secondary diagnostics; the quantitative analysis here centers on faithfulness,
repeatability, and latency, for which complete values are reported.

\section{Full Attack-Followed-by-Repair Results}
\label{app:adversarial}
\newcommand{\EvasionTables}{%
\begin{table}[!htbp]
\centering
\caption{Evasion success (\%) averaged across the four datasets. ``After
repair'' means that FGSM or PGD is optimized first and $\Pi_{\mathcal R}$ is
applied afterward.}
\label{tab:evasion-rates}
\scriptsize
\setlength{\tabcolsep}{3.4pt}
\begin{tabular}{llrrrrr}
\toprule
Attack & Regime & $.01$ & $.05$ & $.10$ & $.20$ & $.30$ \\
\midrule
FGSM & Unrestricted & 20.1 & 46.9 & 63.7 & 74.9 & 76.1 \\
FGSM & After repair & 18.3 & 45.9 & 51.5 & 67.2 & 75.3 \\
PGD & Unrestricted & 21.1 & 54.9 & 74.6 & -- & -- \\
PGD & After repair & 18.5 & 49.1 & 58.8 & -- & -- \\
\bottomrule
\end{tabular}
\end{table}

\begin{table}[!htbp]
\centering
\caption{Normalized trapezoidal area under each evasion-success curve. The
relative change quantifies the observed effect of the repair pipeline.}
\label{tab:evasion-nauc}
\small
\begin{tabular}{lrrr}
\toprule
Attack & Unrestricted & After repair & Relative change \\
\midrule
FGSM & 64.09 & 57.86 & $-9.7\%$ \\
PGD & 52.86 & 44.99 & $-14.9\%$ \\
\bottomrule
\end{tabular}
\end{table}%
}

\newcommand{\CosineTable}{%
\begin{table}[!htbp]
\centering
\caption{Cosine similarity between original and adversarial attribution vectors
under FGSM followed by protocol repair, averaged across datasets. nAUC is the
normalized trapezoidal area over the five displayed budgets.}
\label{tab:cosine-stability}
\scriptsize
\setlength{\tabcolsep}{3.1pt}
\begin{tabular}{lrrrrrr}
\toprule
Method & $.01$ & $.05$ & $.10$ & $.20$ & $.30$ & nAUC \\
\midrule
SHAP & 0.695 & 0.365 & 0.285 & 0.106 & 0.038 & 0.221 \\
LIME & 0.043 & 0.160 & 0.160 & 0.152 & 0.185 & 0.153 \\
IG & 0.706 & 0.374 & 0.316 & 0.080 & -0.025 & 0.212 \\
DeepLIFT & 0.204 & 0.021 & -0.069 & -0.053 & -0.025 & -0.023 \\
PA-SHAP & 0.698 & 0.261 & 0.166 & -0.024 & -0.004 & 0.123 \\
PA-LIME & 0.666 & 0.175 & 0.098 & 0.058 & 0.077 & 0.132 \\
PA-IG & 0.827 & 0.638 & 0.480 & 0.296 & 0.181 & 0.413 \\
PA-DeepLIFT & 0.810 & 0.599 & 0.516 & 0.360 & 0.266 & 0.452 \\
\bottomrule
\end{tabular}
\end{table}%
}

\newcommand{\ScaffoldingTable}{%
\begin{table}[!htbp]
\centering
\caption{Perturbed-feature recovery averaged across datasets. Top-1/Top-3 are
overlaps between the highest-attributed features and features changed by the
attack; lower mean rank is better.}
\label{tab:feature-localization}
\small
\begin{tabular}{lrrr}
\toprule
Method & Top-1 & Top-3 & Mean rank \\
\midrule
LIME & 0.064 & 0.200 & 73.0 \\
SHAP & 0.000 & 0.000 & 123.0 \\
PA-LIME & 0.152 & 0.384 & 41.2 \\
PA-SHAP & 0.038 & 0.122 & 87.5 \\
\bottomrule
\end{tabular}
\end{table}%
}

\subsection{Threat model and evaluated pipeline}
\label{app:attack-math}
The attacked model is the DNN under full-model access. FGSM
\citep{goodfellow2014explaining} is evaluated at
$\varepsilon\in\{0.01,0.05,0.1,0.2,0.3\}$ and PGD
\citep{madry2018towards} at $\varepsilon\in\{0.01,0.05,0.1\}$. The
attack-followed-by-repair composition and its adaptive valid-set extension are
formalized in Appendix~\ref{app:formal-attack}. The curves aggregate outcomes
across the four datasets. Optimizer details, attacked sample indices, and
run-level uncertainty are summarized as additional replication requirements in
Appendix~\ref{app:setup-attacks}.

\subsection{Classifier-evasion curves}
\label{app:evasion-results}
Evasion success is the fraction of initially correct samples that become
misclassified after attack. Table~\ref{tab:evasion-rates} gives the displayed
endpoints. At $\varepsilon=0.1$, repair reduces observed success by $12.2$
percentage points for FGSM and $15.8$ for PGD. For a curve
$(\varepsilon_i,y_i)_{i=1}^{m}$, the normalized trapezoidal summary is
\begin{equation}
 \operatorname{nAUC}(y)=
 \frac{1}{\varepsilon_m-\varepsilon_1}
 \sum_{i=1}^{m-1}
 \frac{y_i+y_{i+1}}{2}
 (\varepsilon_{i+1}-\varepsilon_i).
\label{eq:evasion-nauc}
\end{equation}
Table~\ref{tab:evasion-nauc} shows the corresponding curve averages. These
values characterize the deployed repair pipeline; adaptive constrained
optimization defines a complementary threat model.

\EvasionTables

\begin{figure}[htbp]
  \centering
\begin{minipage}[t]{0.487\linewidth}
\centering
\begin{tikzpicture}
\begin{axis}[
  width=\linewidth,height=4.5cm,
  xmin=0,xmax=0.31,ymin=0,ymax=82,
  xtick={0.0,0.1,0.2,0.3},ytick={0,20,40,60,80},
  xlabel={Attack budget $\varepsilon$},ylabel={Evasion success (\%)},
  title={(a) Classifier evasion},
  tick label style={font=\scriptsize},label style={font=\scriptsize},title style={font=\scriptsize},
  grid=major,grid style={black!8},
  legend style={font=\tiny,at={(0.42,0.58)},anchor=north west,draw=none,row sep=-1pt},
]
\addplot[black,thick,densely dashed,mark=o,mark size=1.7pt,mark options={fill=white}] coordinates {(0.01,20.1) (0.05,46.9) (0.1,63.7) (0.2,74.9) (0.3,76.1)};
\addlegendentry{FGSM before}
\addplot[black,thick,mark=square*,mark size=1.7pt] coordinates {(0.01,18.3) (0.05,45.9) (0.1,51.5) (0.2,67.2) (0.3,75.3)};
\addlegendentry{FGSM after}
\addplot[black!55,thick,densely dashed,mark=triangle,mark size=1.8pt,mark options={fill=white}] coordinates {(0.01,21.1) (0.05,54.9) (0.1,74.6)};
\addlegendentry{PGD before}
\addplot[black!55,thick,mark=diamond*,mark size=1.8pt] coordinates {(0.01,18.5) (0.05,49.1) (0.1,58.8)};
\addlegendentry{PGD after}
\end{axis}
\end{tikzpicture}
\end{minipage}\hfill
\begin{minipage}[t]{0.487\linewidth}
\centering
\begin{tikzpicture}
\begin{axis}[
  width=\linewidth,height=4.5cm,
  xmin=-0.055,xmax=0.48,ymin=0.45,ymax=8.55,y dir=reverse,
  ytick={1,...,8},yticklabels={PA-DeepLIFT,PA-IG,SHAP,IG,LIME,PA-LIME,PA-SHAP,DeepLIFT},xtick={0,0.2,0.4},
  xlabel={Cosine nAUC},title={(b) Curve-summary ranking},
  tick label style={font=\scriptsize},yticklabel style={font=\tiny},label style={font=\scriptsize},title style={font=\scriptsize},
  grid=major,grid style={black!8},clip=false,
]
\addplot[black!55,densely dashed,thin,forget plot] coordinates {(0,0.45) (0,8.55)};
\addplot[black!30,thin,forget plot] coordinates {(0,1) (0.452,1)};
\addplot[black!30,thin,forget plot] coordinates {(0,2) (0.413,2)};
\addplot[black!30,thin,forget plot] coordinates {(0,3) (0.221,3)};
\addplot[black!30,thin,forget plot] coordinates {(0,4) (0.212,4)};
\addplot[black!30,thin,forget plot] coordinates {(0,5) (0.153,5)};
\addplot[black!30,thin,forget plot] coordinates {(0,6) (0.132,6)};
\addplot[black!30,thin,forget plot] coordinates {(0,7) (0.123,7)};
\addplot[black!30,thin,forget plot] coordinates {(0,8) (-0.023,8)};
\addplot[only marks,mark=o,mark size=2pt,very thick,black,mark options={fill=white}] coordinates {(0.221,3) (0.212,4) (0.153,5) (-0.023,8)};
\addplot[only marks,mark=square*,mark size=1.9pt,black] coordinates {(0.452,1) (0.413,2) (0.132,6) (0.123,7)};
\end{axis}
\end{tikzpicture}
\end{minipage}
  \caption{Module~2 summaries. (a) Mean classifier-evasion success before and
  after protocol repair; PGD records end at $\varepsilon=0.1$. (b) Normalized
  trapezoidal area of the five-point cosine curves under FGSM followed by
  repair; higher values indicate greater average directional alignment. Open
  circles denote standard methods and filled squares PA methods.}
  \label{fig:appendix-module2-summary}
\end{figure}
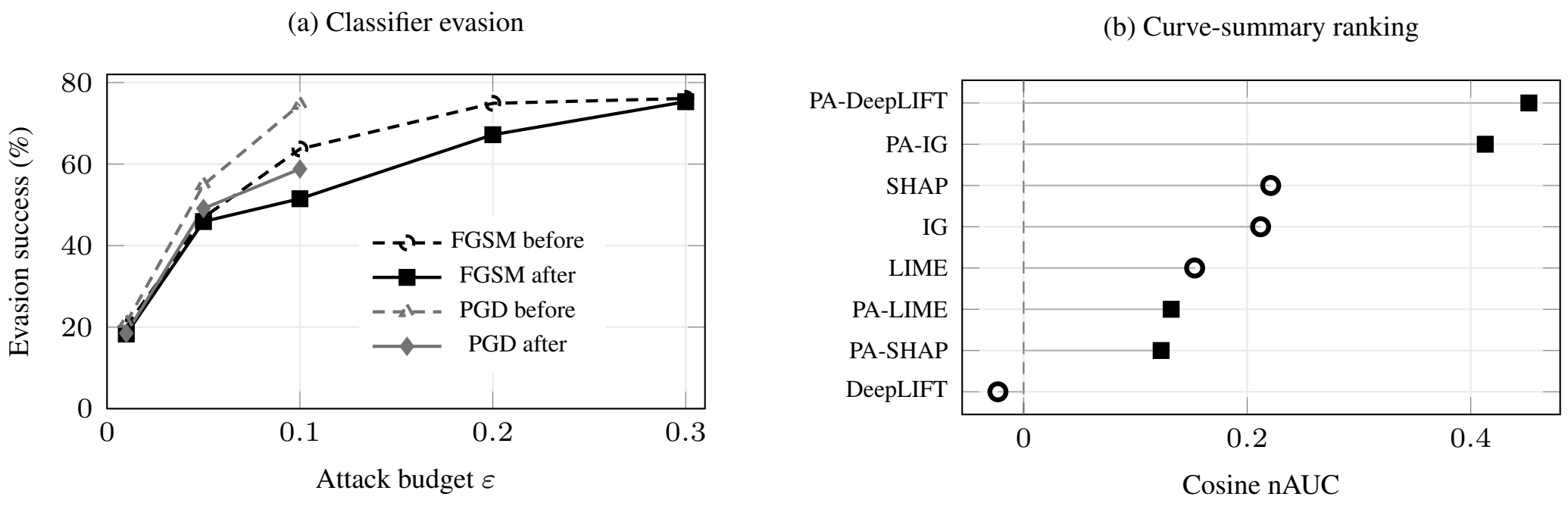

\subsection{Attribution similarity under repaired FGSM}
\label{app:attribution-stability}
Cosine similarity compares the attribution vector at the original input with
the vector at the repaired attack output. Higher values indicate greater
directional alignment, while negative values indicate opposing directions.
Table~\ref{tab:cosine-stability} gives all displayed values and their nAUC from
Equation~\eqref{eq:evasion-nauc}. PA-DeepLIFT and PA-IG have the two largest
curve summaries, $0.452$ and $0.413$. The ordering varies by budget, and the
PA-LIME and PA-SHAP curve summaries remain below their standard counterparts.

Three additional reported point values complement the curves: PA-DeepLIFT
class-equivalent Jaccard is $0.868$ at $\varepsilon=0.01$ and $0.608$ at
$\varepsilon=0.1$; the LIME Lipschitz ratio exceeds $15$ at
$\varepsilon=0.01$; and SHAP cosine similarity at $\varepsilon=0.1$ is $0.171$
before repair and $0.285$ after repair. Complete arrays for these secondary
metrics would enable curve-level comparison.

\CosineTable

\subsection{Perturbed-feature localization}
\label{app:scaffolding}
Table~\ref{tab:feature-localization} measures whether highly attributed
features coincide with features changed by the attack. Relative to LIME,
PA-LIME raises Top-1 overlap from $0.064$ to $0.152$ and Top-3 overlap from
$0.200$ to $0.384$, while reducing mean rank from $73.0$ to $41.2$. These are
$2.38\times$, $1.92\times$, and $43.6\%$ improvements, respectively. The
absolute rates position the result as a useful investigative cue rather than an
automated response mechanism.

\ScaffoldingTable

\subsection{Threat-model interpretation}
\label{app:adversarial-interpretation}
Module~2 establishes how the evaluated repair pipeline changes classifier
evasion, attribution similarity, and perturbed-feature localization for the
specified DNN, budgets, datasets, and explainers. Adaptive valid-set attacks,
direct explanation-manipulation objectives, repair-aware norm control,
support-consistency tests, raw-flow reconstruction, and analyst decision studies
provide complementary extensions of this threat model.

\section{Full Agreement, Interaction, and Reference-Set Results}
\label{app:consensus}
\subsection{Agreement protocol and aggregate results}
The aggregate pool comprises 66 configuration pairs per dataset, consistent
with 12 evaluated configurations, and uses $10{,}000$ samples per dataset.
Agreement is summarized by Spearman $\rho_s$, Kendall $\tau_b$, Top-5 and
Top-10 overlap, and a Wilcoxon signed-rank test at nominal $\alpha=0.05$.
Publishing the exact configuration and pair manifest would enable a complete
factor-level replication.

Table~\ref{tab:consensus-full} reports the complete dataset-level agreement
summary, including rank correlations, top-$k$ overlap, Wilcoxon rejection
rates, and within-mode means.

\FullConsensusTable

All-pair mean Spearman agreement is $0.256$--$0.309$, Kendall agreement is
$0.196$--$0.249$, and Top-5 overlap is $0.168$--$0.322$. The Wilcoxon rejection
share is $84.8\%$--$89.4\%$. Within-PA Spearman means exceed within-standard
means on all four datasets; their four-dataset averages are $0.366$ and $0.217$,
for a difference of $0.149$. Together, these values show stronger within-PA
coherence against a background of substantial configuration-level variation.

\subsection{Pair examples and the model--explainer structure}
\label{app:consensus-confound}
The aggregate analysis characterizes configuration-level variability.
Fixed-model examples isolate explainer disagreement, whereas cross-model pairs
capture the combined effect of model and explanation choice.

Table~\ref{tab:consensus-pair-examples} makes this distinction explicit by
separating fixed-model explainer contrasts from pairs that also change the
model family or generation mode.

\PairExamplesTable

DNN-LIME versus DNN-SHAP holds the predictive model fixed and shows explainer
disagreement on NSL-KDD. PA-DeepLIFT and PA-IG share the DNN and the
threshold-near reference rule, making their high correlations evidence of
configuration coherence rather than an independent correctness check. The
DNN-PA-LIME versus RF-LIME example changes both the model and generation mode.

\subsection{Reported interaction statistics}
\label{app:interaction-results}
Table~\ref{tab:interaction-safe} reports the model-specific interaction
statistics and their DNN/RF quotients, obtained by dividing the displayed
DNN value by the RF value and rounding to two decimals. The sample and
feature-pair aggregation and normalization of the base statistics are
unspecified; the quotients therefore compare reported values rather than a
defined cross-model interaction-strength measure.

\InteractionTable

The DNN statistic exceeds the RF statistic on each dataset, and the quotients
range from $1.02$ to $1.60$. Two selected feature examples show large ranking
discrepancies: \textit{Protocol} on UNSW-NB15 differs by 71 positions between
LIME and the interaction ranking, and \textit{Bwd Pkt Len Std} on
CSE-CIC-IDS2018 differs by 70 positions. Interpreting these values as
interaction strength requires the missing aggregation and normalization rules.

\subsection{Predefined domain feature sets}
\label{app:reference-sets}
For attack type $a$, let $S_a$ denote a predefined domain feature set and
$\operatorname{Top}_{15}(\phi)$ the 15 highest-ranked features of explanation
$\phi$. The reported measure is
\begin{equation}
 \operatorname{RRA@15}(a,\phi)
 =\frac{|S_a\cap\operatorname{Top}_{15}(\phi)|}{|S_a|}.
 \label{eq:rra15}
\end{equation}
Equation~\eqref{eq:rra15} measures coverage of the predefined attack-domain
set, whose denominator is the set size. The predefined sets contain 6--8
attack-relevant features. Publishing their
exact membership and construction protocol would enable independent replication
of RRA@15.

Table~\ref{tab:expert-reference-selected} summarizes the retained
attack-specific RRA@15 leaders and the reported all-attack PA-LIME average.
The changing leaders across attack types reinforce that domain-set alignment
is configuration-specific.

\ReferenceSetTable

The reported leaders vary by attack: DeepLIFT leads the listed Bot and
FTP-BruteForce cases, PA-IG leads the three listed DoS/Infiltration cases, and
PA-SHAP leads the SSH-Bruteforce and DDoS-LOIC-HTTP cases. PA-LIME has a
reported all-attack mean of $0.064$. RRA@15 asks whether the top 15
attributions recover predefined attack-domain features, while Module~2
localization asks whether highly ranked features were changed by the evaluated
perturbation (Table~\ref{tab:feature-localization}). These targets and
aggregation scopes differ, so higher perturbation localization can coexist
with lower domain-set recovery.

\subsection{Interpretation}
Agreement, interaction rankings, and domain-reference alignment provide
complementary corroboration across configurations. Their scientific role is to
identify consistency, disagreement, and domain-signature correspondence, while
Module~1 anchors the audit in model-response alignment.

\section{Full Attribution-Guided Feature-Reduction Results}
\label{app:reduction}
\newcommand{\FullReductionTable}{%
\par\medskip
\refstepcounter{table}\label{tab:module4-full-coordinates}%
\noindent\begin{minipage}{\linewidth}
\centering
\footnotesize
\textbf{Table \thetable:} All 16 aggregate coordinates. Higher compression and
a higher (less negative) $\Delta F1_w$ are preferred. ``ND'' marks
nondominance among the displayed points.\\[0.45em]
\setlength{\tabcolsep}{3.8pt}
\begin{tabular}{@{}llrrc@{}}
\toprule
Family & Configuration & Compression (\%) & $\Delta F1_w$ (pp) & ND \\
\midrule
Conventional & Chi-Squared & 80.6 & -0.87 & no \\
Conventional & PCA & 80.6 & -0.69 & no \\
Conventional & Spearman & 80.6 & -0.70 & no \\
Conventional & InfoGain & 80.6 & -0.80 & no \\
\addlinespace[1pt]
Standard XAI & SHAP-DNN & 86.5 & -1.88 & no \\
Standard XAI & SHAP-RF & 82.2 & -1.15 & no \\
Standard XAI & LIME-DNN & 59.1 & -1.15 & no \\
Standard XAI & LIME-RF & 54.4 & -1.04 & no \\
Standard XAI & IG-DNN & 88.5 & -2.25 & no \\
Standard XAI & DeepLIFT-DNN & 84.5 & -2.25 & no \\
\addlinespace[1pt]
PA-XAI & PA-SHAP-DNN & 85.3 & -0.68 & \textbf{yes} \\
PA-XAI & PA-SHAP-RF & 85.1 & -1.48 & no \\
PA-XAI & PA-LIME-DNN & 84.3 & -4.27 & no \\
PA-XAI & PA-LIME-RF & 86.0 & -2.91 & no \\
PA-XAI & PA-IG-DNN & 90.5 & -0.89 & \textbf{yes} \\
PA-XAI & PA-DeepLIFT-DNN & 89.6 & -0.94 & no \\
\bottomrule
\end{tabular}
\end{minipage}
\par\medskip
}

\newcommand{\PCAComparisonTable}{%
\par\medskip
\refstepcounter{table}\label{tab:module4-pca-comparison}%
\noindent\begin{minipage}{\linewidth}
\centering
\small
\textbf{Table \thetable:} Selected PA coordinates relative to the PCA coordinate
$(80.6,-0.69)$. Differences are PA minus PCA in percentage points.\\[0.45em]
\setlength{\tabcolsep}{5pt}
\begin{tabular}{@{}lrrrr@{}}
\toprule
Configuration & Compression & $\Delta F1_w$ & $\Delta$ compression & $\Delta\Delta F1_w$ \\
\midrule
PA-SHAP-DNN & 85.3 & -0.68 & +4.7 & +0.01 \\
PA-IG-DNN & 90.5 & -0.89 & +9.9 & -0.20 \\
PA-DeepLIFT-DNN & 89.6 & -0.94 & +9.0 & -0.25 \\
\bottomrule
\end{tabular}
\end{minipage}
\par\medskip
}

\subsection{Procedure and metric scope}
\label{app:reduction-procedure}
Module~4 supplies the utility coordinate $U$. A global feature ranking is
computed from mean absolute attribution over $5{,}000$ calibration samples. A
separate RF proxy with 50 estimators and maximum depth 10, trained on a
$50{,}000$-sample subsample, prunes features until macro-F1 decreases by more
than 0.01 from the full-feature proxy baseline. A DNN, 1D-CNN, RF, and SVM are
then retrained on the selected subset; the same procedure is applied to
chi-squared, Information Gain, PCA, and Spearman baselines.

The horizontal coordinate is the reported percentage reduction in feature
dimension. The vertical coordinate is the median matched-classifier
weighted-F1 change, in percentage points, relative to the corresponding
full-feature classifier. Aggregate results are reported over the available
matched classifier--dataset evaluations.

\subsection{Displayed coordinates and nondominance}
\label{app:reduction-frontier}
For displayed points $a=(c_a,d_a)$ and $b=(c_b,d_b)$, where higher compression
$c$ and higher weighted-F1 change $d$ are preferred, $a$ dominates $b$ when
\begin{equation}
 c_a\ge c_b,\qquad d_a\ge d_b,
 \qquad\text{and at least one inequality is strict.}
 \label{eq:reduction-dominance}
\end{equation}
A point is nondominated when no other displayed point dominates it. Applying
this rule to the 16 aggregate coordinates yields two nondominated points:
PA-SHAP-DNN and PA-IG-DNN.

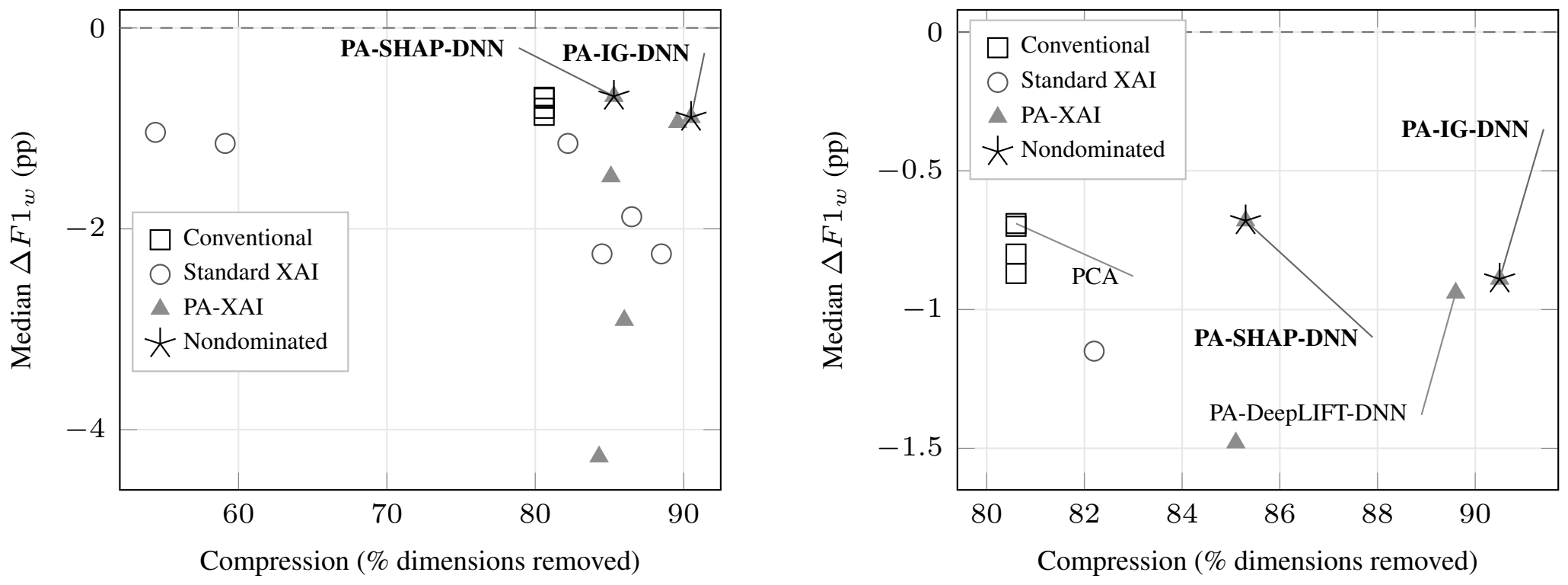
\begin{figure}[t]
  \centering
\begin{minipage}[t]{0.487\linewidth}
\centering
\begin{tikzpicture}
\begin{axis}[
  width=\linewidth,height=5.75cm,
  xmin=52,xmax=92.5,ymin=-4.6,ymax=0.18,
  xlabel={Compression (\% dimensions removed)},ylabel={Median $\Delta F1_w$ (pp)},
  tick label style={font=\scriptsize},label style={font=\scriptsize},
  grid=major,grid style={black!9},clip=false,
  legend style={font=\tiny,at={(0.02,0.58)},anchor=north west,draw=black!25,fill=white,row sep=-1pt},
  legend cell align={left},
]
\addplot[black!55,densely dashed,thin,forget plot] coordinates {(52,0) (92.5,0)};
\addplot[only marks,mark=square,mark size=2.4pt,black] coordinates {(80.6,-0.87) (80.6,-0.69) (80.6,-0.7) (80.6,-0.8)};
\addlegendentry{Conventional}
\addplot[only marks,mark=o,mark size=2.4pt,black!65] coordinates {(86.5,-1.88) (82.2,-1.15) (59.1,-1.15) (54.4,-1.04) (88.5,-2.25) (84.5,-2.25)};
\addlegendentry{Standard XAI}
\addplot[only marks,mark=triangle*,mark size=2.4pt,black!45] coordinates {(85.3,-0.68) (85.1,-1.48) (84.3,-4.27) (86,-2.91) (90.5,-0.89) (89.6,-0.94)};
\addlegendentry{PA-XAI}
\addplot[only marks,mark=star,mark size=4.0pt,black] coordinates {(85.3,-0.68) (90.5,-0.89)};
\addlegendentry{Nondominated}

\draw[black!60,thin] (axis cs:85.3,-0.68) -- (axis cs:78.9,-0.20);
\node[font=\tiny\bfseries,anchor=east] at (axis cs:78.9,-0.20) {PA-SHAP-DNN};
\draw[black!60,thin] (axis cs:90.5,-0.89) -- (axis cs:91.4,-0.25);
\node[font=\tiny\bfseries,anchor=east] at (axis cs:91.4,-0.25) {PA-IG-DNN};

\end{axis}
\end{tikzpicture}
\end{minipage}\hfill
\begin{minipage}[t]{0.487\linewidth}
\centering
\begin{tikzpicture}
\begin{axis}[
  width=\linewidth,height=5.75cm,
  xmin=79.4,xmax=91.7,ymin=-1.65,ymax=0.08,
  xlabel={Compression (\% dimensions removed)},ylabel={Median $\Delta F1_w$ (pp)},
  tick label style={font=\scriptsize},label style={font=\scriptsize},
  grid=major,grid style={black!9},clip=false,
  legend style={font=\tiny,at={(0.02,0.98)},anchor=north west,draw=black!25,fill=white,row sep=-1pt},
  legend cell align={left},
]
\addplot[black!55,densely dashed,thin,forget plot] coordinates {(79.4,0) (91.7,0)};
\addplot[only marks,mark=square,mark size=2.4pt,black] coordinates {(80.6,-0.87) (80.6,-0.69) (80.6,-0.7) (80.6,-0.8)};
\addlegendentry{Conventional}
\addplot[only marks,mark=o,mark size=2.4pt,black!65] coordinates {(82.2,-1.15)};
\addlegendentry{Standard XAI}
\addplot[only marks,mark=triangle*,mark size=2.4pt,black!45] coordinates {(85.3,-0.68) (85.1,-1.48) (90.5,-0.89) (89.6,-0.94)};
\addlegendentry{PA-XAI}
\addplot[only marks,mark=star,mark size=4.0pt,black] coordinates {(85.3,-0.68) (90.5,-0.89)};
\addlegendentry{Nondominated}

\draw[black!60,thin] (axis cs:85.3,-0.68) -- (axis cs:87.9,-1.10);
\node[font=\tiny\bfseries,anchor=east] at (axis cs:87.9,-1.10) {PA-SHAP-DNN};
\draw[black!60,thin] (axis cs:90.5,-0.89) -- (axis cs:91.4,-0.35);
\node[font=\tiny\bfseries,anchor=east] at (axis cs:91.4,-0.35) {PA-IG-DNN};

\draw[black!45,thin] (axis cs:80.6,-0.69) -- (axis cs:83.0,-0.88);
\node[font=\tiny,anchor=east] at (axis cs:83.0,-0.88) {PCA};
\draw[black!45,thin] (axis cs:89.6,-0.94) -- (axis cs:88.9,-1.38);
\node[font=\tiny,anchor=east] at (axis cs:88.9,-1.38) {PA-DeepLIFT-DNN};

\end{axis}
\end{tikzpicture}
\end{minipage}
  \caption{All 16 compression--fidelity coordinates (left) and an enlarged view
  of the frontier region (right). Rightward and upward are preferred; the dashed
  line marks no weighted-F1 change, and stars identify the two nondominated
  points.}
  \label{fig:appendix-module4-frontier}
\end{figure}

Figure~\ref{fig:appendix-module4-frontier} visualizes the complete
16-configuration set and enlarges the frontier region.
Table~\ref{tab:module4-full-coordinates} reports the exact coordinates and
confirms that only PA-SHAP-DNN and PA-IG-DNN are nondominated.

\FullReductionTable

\subsection{Comparison with the conventional operating point}
\label{app:reduction-comparisons}
At the reported operating point, the four conventional methods cluster at
$80.6\%$ compression with median weighted-F1 changes from $-0.87$ to $-0.69$
percentage points. PCA has the strongest displayed conventional coordinate,
$(80.6,-0.69)$.

Table~\ref{tab:module4-pca-comparison} quantifies the additional compression
and weighted-F1 trade-offs of the selected PA configurations relative to PCA.

\PCAComparisonTable

PA-SHAP-DNN adds $4.7$ compression points relative to PCA and has a
$+0.01$-point weighted-F1 difference. PA-IG-DNN adds $9.9$ compression points
with a $-0.20$-point difference. PA-DeepLIFT-DNN adds $9.0$ compression points
with a $-0.25$-point difference and is dominated by PA-IG-DNN. PCA supplies a
transformed-component compression benchmark, while the attribution methods
retain original-feature semantics.

\subsection{Method-specific trade-offs and auxiliary results}
\label{app:reduction-tradeoffs}
Standard LIME-DNN and LIME-RF report $59.1\%$ and $54.4\%$ compression with
weighted-F1 changes of $-1.15$ and $-1.04$ points; their mean retained-feature
counts are 42.7 and 46.3. PA-LIME-DNN and PA-LIME-RF move to more aggressive
compression points, $84.3\%$ and $86.0\%$, with larger fidelity costs of
$-4.27$ and $-2.91$ points. PA-LIME-DNN has a reported IQR of $8.12$
percentage points. These values identify a distinct sparsity--fidelity operating
regime for PA-LIME.

Selected PA-IG-DNN runs retain 6--8 reported features on the complex datasets
considered. Per-dataset feature counts and selected feature lists would refine
this aggregate observation.

\subsection{Scope of the compression--fidelity result}
The displayed frontier shows that selected attribution rankings preserve
predictive utility under strong compression, with PA-SHAP-DNN and PA-IG-DNN
nondominated among the 16 aggregate configurations. Per-dataset frontiers,
deployment cost, and adversarial behavior of the reduced models are separate
evaluation targets.

\section{Derivation of Aggregate Summaries}
\label{app:numerical}

This appendix gives the arithmetic behind the aggregate summaries reported in
the main paper.

Table~\ref{tab:numerical-derivations} links each derived headline result to
its displayed inputs, including the stability mean, sign-change count,
agreement difference, nAUC summaries, localization ratios, and Pareto result.

\begin{table}[htbp]
\centering
\caption{Derivation of aggregate results from the displayed values.}
\label{tab:numerical-derivations}
\footnotesize
\setlength{\tabcolsep}{3pt}
\begin{tabular}{@{}p{0.31\linewidth}p{0.43\linewidth}p{0.17\linewidth}@{}}
\toprule
\textbf{Result} & \textbf{Calculation} & \textbf{Interpretation} \\
\midrule
Six-case PA-LIME mean & $\operatorname{mean}(0.48,0.39,0.69,0.86,0.43,0.60)=0.575$ & Paired stability summary \\
Stability difference & $0.575-1.000=-0.425$ & Change from standard perfect overlap \\
Faithfulness sign changes & $5/12$ comparisons; DNN $4/4$ & Qualitative evaluator changes \\
Within-mode agreement & PA $0.366$, standard $0.217$, difference $0.149$ & Descriptive coherence gap \\
FGSM evasion nAUC & $64.09\rightarrow57.86$; $-9.7\%$ & Repair-pipeline effect \\
PGD evasion nAUC & $52.86\rightarrow44.99$; $-14.9\%$ & Repair-pipeline effect \\
Cosine nAUC leaders & PA-DeepLIFT $0.452$; PA-IG $0.413$ & Curve-level ranking \\
PA-LIME localization & Top-1 $2.38\times$; Top-3 $1.92\times$; rank $-43.6\%$ & Improved investigative cue \\
Displayed Pareto set & PA-SHAP-DNN; PA-IG-DNN & Nondominated aggregate points \\
\bottomrule
\end{tabular}
\end{table}

For a curve $(x_i,y_i)_{i=1}^{m}$, nAUC is the trapezoidal area normalized by
the observed horizontal span:
\begin{equation}
\operatorname{nAUC}(y)=
\frac{1}{x_m-x_1}
\sum_{i=1}^{m-1}
\frac{y_i+y_{i+1}}{2}(x_{i+1}-x_i).\label{eq:aggregate-nauc}
\end{equation}
Equation~\eqref{eq:aggregate-nauc} averages curve height over the observed
budget span; it introduces no unobserved endpoint. The Pareto result follows from the dominance relation in
Equation~\eqref{eq:reduction-dominance} applied to all 16 displayed
compression--fidelity coordinates.

\end{document}